%% file: bare_jrnl_compsoc.tex
\documentclass[10pt,journal]{IEEEtran}

\usepackage{amsmath,amssymb,amsfonts}
\usepackage{graphicx}
\usepackage{textcomp}
\usepackage{xcolor}
\usepackage{algorithm}
\usepackage{algpseudocode}
\usepackage{mathrsfs}
\usepackage{overpic}
\usepackage{tikz}
\usepackage{cite}
\usepackage{array}
\usepackage{etoolbox}
\usepackage{caption} 
\usepackage[caption=false,font=footnotesize,labelfont=sf,textfont=sf]{subfig}
\usepackage{stfloats}
\usepackage{url}
\usepackage{multirow}
\usepackage{booktabs}

\newcommand{\insertfig}{\setcounter{figure}{0}\includegraphics[width= 0.96\linewidth]{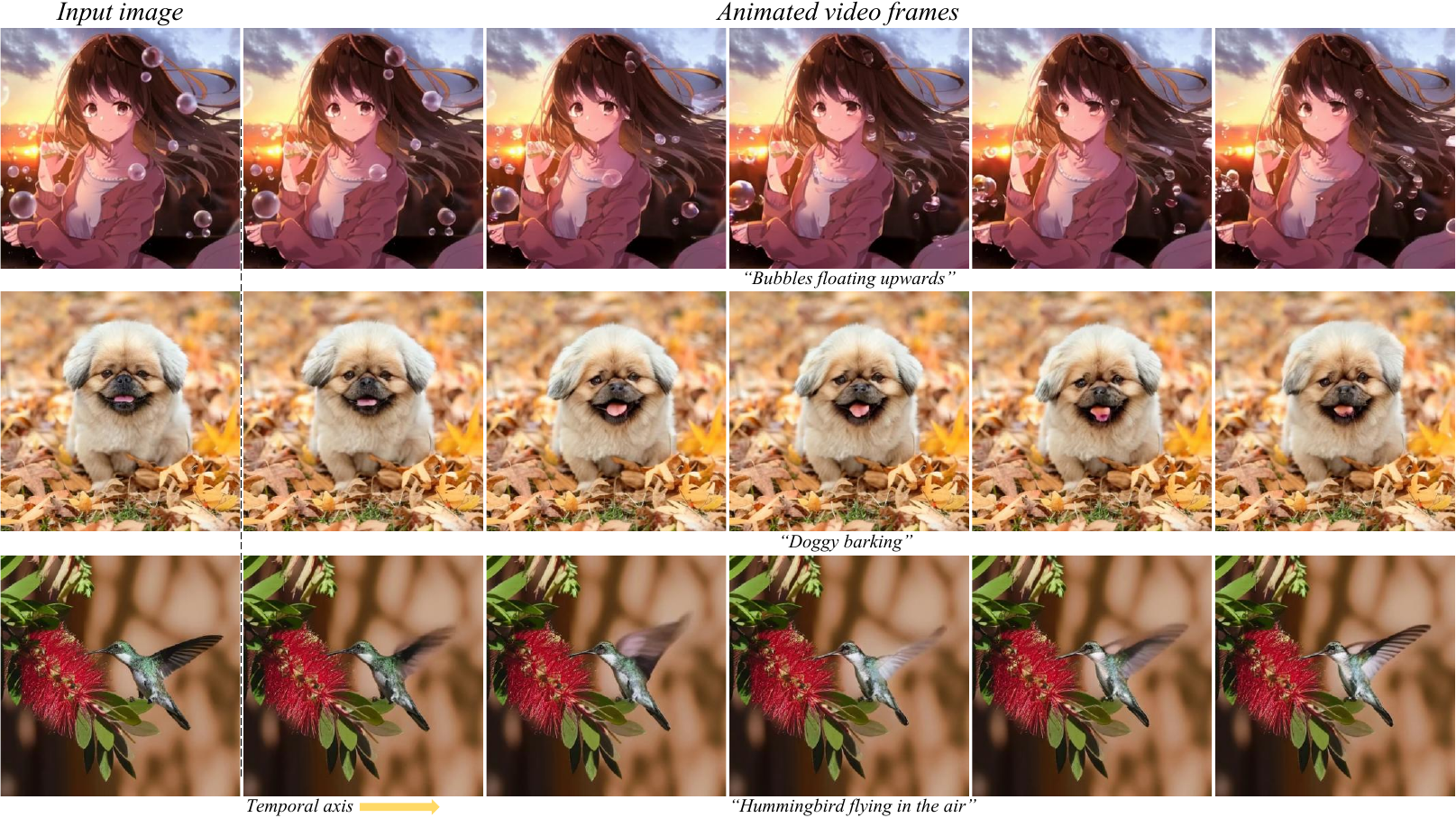}\captionof{figure}{Image animation from our method. Please visit the project page to visualize the animations.}\label{fig_image_animation_teaser}}

\makeatletter
\apptocmd{\@maketitle}{\centering \insertfig}{}{}
\makeatother

\begin{document}
%
\title{Consistent and Controllable Image Animation with Motion Linear Diffusion Transformers}

\author{Xin~Ma,
        Yaohui~Wang$\ddagger$,
        Genyun~Jia,
        Xinyuan~Chen,
        Tien-Tsin~Wong,~\IEEEmembership{Member,~IEEE}
        and~Cunjian~Chen$\ddagger$,~\IEEEmembership{Senior~Member,~IEEE}
\IEEEcompsocitemizethanks{\IEEEcompsocthanksitem Xin Ma, Tien-Tsin Wong and Cunjian Chen are with the Department of Data Science \& AI, Faculty of Information Technology, Monash
 University, Australia,
Melbourne, 3800. E-mail: \{xin.ma1, tt.wong, cunjian.chen\}@monash.edu
\IEEEcompsocthanksitem Genyun Jia is with Nanjing University of Posts and Telecommunications, China, Nanjing, 210003. E-mail: gengyun.jia@njupt.edu.cn
\IEEEcompsocthanksitem Yaohui Wand and Xinyuan Chen are with Shanghai Artificial Intelligence Laboratory, China, Shanghai, 200240. E-mail: \{wangyaohui, chenxinyuan\}@pjlab.org.cn
}
\thanks{$\ddagger$ Corresponding authors.}
}


\IEEEtitleabstractindextext{%
\begin{abstract}
Image animation has seen significant progress, driven by the powerful generative capabilities of diffusion models.
However, maintaining appearance consistency with static input images and mitigating abrupt motion transitions in generated animations remain persistent challenges. While text-to-video (T2V) generation has demonstrated impressive performance with diffusion transformer models, the image animation field still largely relies on U-Net-based diffusion models, which lag behind the latest T2V approaches. Moreover, the quadratic complexity of vanilla self-attention mechanisms in Transformers imposes heavy computational demands, making image animation particularly resource-intensive.
To address these issues, we propose MiraMo, a framework designed to enhance efficiency, appearance consistency, and motion smoothness in image animation.
Specifically, MiraMo introduces three key elements: (1) A foundational text-to-video architecture replacing vanilla self-attention with efficient linear attention to reduce computational overhead while preserving generation quality; (2) A novel motion residual learning paradigm that focuses on modeling motion dynamics rather than directly predicting frames, improving temporal consistency; and (3) A DCT-based noise refinement strategy during inference to suppress sudden motion artifacts, complemented by a dynamics control module to balance motion smoothness and expressiveness.
Extensive experiments against state-of-the-art methods validate the superiority of MiraMo in generating consistent, smooth, and controllable animations with accelerated inference speed. Additionally, we demonstrate the versatility of MiraMo through applications in motion transfer and video editing tasks. The project page is available at~\url{https://maxin-cn.github.io/miramo_project}.
\end{abstract}

}

\maketitle

\IEEEdisplaynontitleabstractindextext

%
\IEEEpeerreviewmaketitle

%
%
%
%

\input{sections/introduction}
\input{sections/related}
\input{sections/method}
\input{sections/experiment}
\input{sections/conclusion}



\bibliographystyle{IEEEtran}
\bibliography{bibliography}

\ifCLASSOPTIONcaptionsoff
  \newpage
\fi




\end{document}

%% file: sections/introduction.tex
\section{Introduction}
\label{sec:introduction}

\begin{figure*}
\centering
\includegraphics[width=1.0\linewidth]{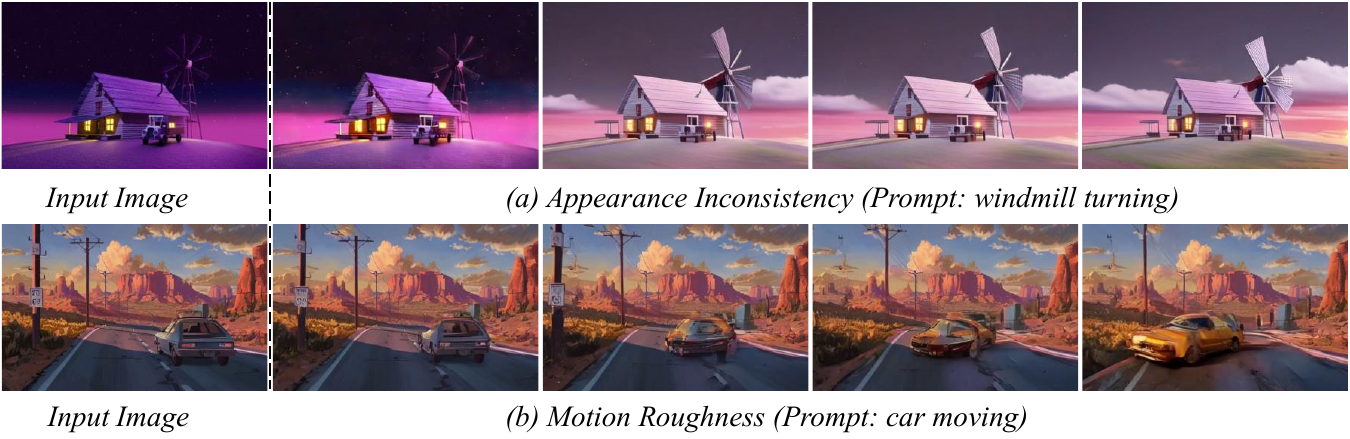}
\caption{\textbf{Appearance consistency and motion smoothness.} 
(a) Upper row: appearance changes over time (result from PIA~\cite{zhang2023pia}). 
(b) Lower row: the car takes a sudden turnaround (result from SEINE~\cite{chen2023seine}).}
\label{fig_existing_methods_issues}
\end{figure*}


\IEEEPARstart{I}{mage} animation, also known as Image-to-Video generation (I2V), ``animates'' a given static image to transform it into a video sequence, with the goal of preserving the original content but injecting realistic and temporally coherent dynamics. Fig.~\ref{fig_image_animation_teaser} demonstrates some examples (please visit the project page to visualize the motion). It is an engaging and convenient way of creating video content, which can be directly used in real-world video production such as photography, filmmaking, and augmented reality. 

Recent advancements in large-scale diffusion models have enabled generalization to open domains, demonstrating success in image \cite{rombach2022high,chen2023pixart,podell2023sdxl}, 3D geometry \cite{lin2023magic3d,poole2022dreamfusion,li2023instant3d}, and video generation \cite{wang2023lavie,ma2024latte,blattmann2023align,luo2023videofusion,singer2022make,guo2024animatediff}. In particular, efforts to add an additional image condition to text-to-video (T2V) diffusion models, thus enabling image animation through the leveraging of their powerful spatio-temporal generation priors, have attracted much attention~\cite{zhang2023pia,dai2023animateanything,xing2023dynamicrafter,zhang2023i2vgen,ren2024consisti2v,chen2023seine,shi2025motionstone,Yariv_2025_CVPR,tian2025extrapolating,gui2025i2vguard,zhang2025motionpro,wang2025levitor,mao2025osv,wang2025motif}.
Despite these advancements, as demonstrated in Fig.~\ref{fig_existing_methods_issues}, existing image animation methods still encounter two major challenges: \emph{appearance consistency} and \emph{motion smoothness}. Firstly, the generated videos may fail to maintain the visual appearance over time (the overall color tone and the shape of the windmill in Fig.~\ref{fig_existing_methods_issues}(a)). Secondly, abrupt motion or deformation of objects may happen as shown in Fig.~\ref{fig_existing_methods_issues}(b). Meanwhile, recent T2V models have largely adopted Transformer-based architectures due to their superior performance over U-Net-based models~\cite{kong2024hunyuanvideo,wan2025wan,yang2024cogvideox,lin2024open,zheng2024open,hacohen2024ltx}. In contrast, the image animation field still predominantly relies on U-Net-based diffusion models, which lag behind the latest T2V methods~\cite{wang2025motif,wang2025levitor,tian2025extrapolating}. However, a commonly acknowledged drawback of Transformers is that the standard quadratic attention mechanisms significantly increase both training and inference efficiency.

A prevailing approach to ensuring appearance consistency and motion smoothness is to constrain the motion dynamic degree. Yet ideally, one would preserve such consistencies without compromising on substantial dynamic motion, a balance that remains an enduring challenge in image animation. In this paper, we aim to strike such a balance by proposing two novel designs in our model, \textbf{MiraMo}, which is based on a self-constructed and fast T2V model with a Linear Transformer.
Firstly, we design a foundational T2V model that replaces all vanilla attention modules in the Transformer block with more efficient linear attention mechanisms, which reduces the computational complexity from $O(N^2)$ to $O(N)$. To enable our linear Transformer to capture temporal relationships in videos, we introduce a different linear attention mechanism that is well compatible with RoPE positional encoding~\cite{rope}. During the post-training stage of the T2V model, we find that using a synthetic video dataset can significantly improve the quality of generated videos compared to a carefully curated small-scale high-quality real-world dataset. We believe that linear attention can achieve performance comparable to vanilla attention, and with appropriate design, it can generate videos more efficiently.
Secondly, to better preserve the appearance consistency across frames, we propose to learn the distribution of the differences ({\em motion residual}) between frames based on the above self-constructed T2V model, instead of direct learning on the multiple frames themselves. 
Finally, we propose to utilize the low-frequency Discrete Cosine Transform (DCT) components of the input image as the layout guidance to suppress the abrupt or undesired motion during the inference phase. Such process ({\em DCTInit})
is able to address the noise discrepancy issue between training and inference phases as mentioned in other works~\cite{si2024freeu, wu2023freeinit, ning2023input} and is effective in suppressing the abrupt motion as demonstrated in Fig.~\ref{fig_fft_vs_dct_e_dctinit}(c\&d).
To avoid over-suppression of motion dynamics, we propose a new design of {\em dynamics degree control} inspired by~\cite{dai2023animateanything} that enables users to explicitly control the degree of motion. 

With the proposed strategies, our model is able to generate videos with high appearance and motion consistency that accurately align with the input prompts, all while maintaining fast inference speed. We also demonstrate that our model can be easily extended to other applications such as video editing and motion transfer. We conduct comprehensive quantitative and qualitative experiments and demonstrate that our model achieves the best appearance consistency and motion smoothness, without over-suppressing the motion dynamics. We summarize our contributions as follows:

\begin{itemize}
\item We propose MiraMo, a flow matching based image animation method that enables fast inference through the use of a linear Transformer.

\item A different linear attention mechanism, compatible with Rotary Positional Embedding (RoPE) positional embedding, is introduced to effectively capture temporal relationships in videos.

\item We find that using a synthetic video dataset can significantly improve the quality of generated videos compared to using a carefully curated small-scale high-quality real-world dataset during the post-training stage.

\item Based on our self-constructed T2V model, MiraMo learns the distribution of motion residuals across frames rather than directly predicting the multiple frames themselves for image animation, in turn better preserving the appearance consistency.

\item To mitigate the abrupt or undesired motion changes, we utilize the low-frequency DCT component of the input image to initialize the inference noise during the inference phase. 
\end{itemize}

This paper extends and improves upon the conference paper \textit{``Consistent and Controllable Image Animation with Motion Diffusion Models''} published in CVPR 2025~\cite{ma2025consistent} in the following aspects. 
Firstly, while Transformer-based models have been proven superior to U-Net architectures in the T2V domain, U-Net-based video generation models are still widely adopted in image animation, leaving the field significantly behind recent advances. To bridge this gap, we explore Transformer-based approaches for image animation and adopt a flow matching loss function. 
Secondly, given the quadratic computational complexity $O(N^2)$ of the vanilla attention mechanism in Transformers and inspired by the success of linear attention in text-to-image generation (T2I)~\cite{xie2024sana}, we introduce a novel linear Transformer for video generation that reduces the complexity from $O(N^2)$ to $O(N)$. As shown in Fig.~\ref{fig:throughput_gpu}, our MiraMo achieves higher throughput, lower GPU memory consumption, and better performance compared to the vanilla attention-based Transformer model.
Thirdly, to enable our linear Transformer to effectively capture temporal relationships in videos, we introduce a new linear attention mechanism that is compatible with RoPE positional embedding. 
Fourthly, to further improve the final video quality of the linear Transformer-based model, we investigate how the choice of video data impacts performance during the post-training stage. 
Finally, building on our self-developed T2V model, we construct an image animation model that achieves significantly faster generation without noticeable loss in generation quality.

%% file: sections/related.tex
\section{Related Work}

\begin{figure}[ht]  
\centering
\subfloat
  {
    \includegraphics[scale=0.5]{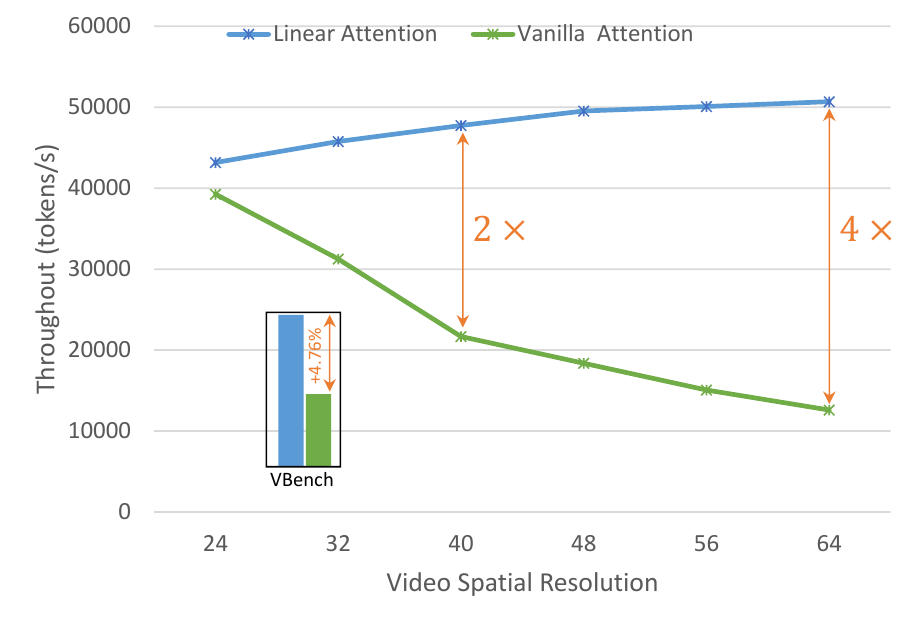}
  } \\
\subfloat
  {
    \includegraphics[scale=0.5]{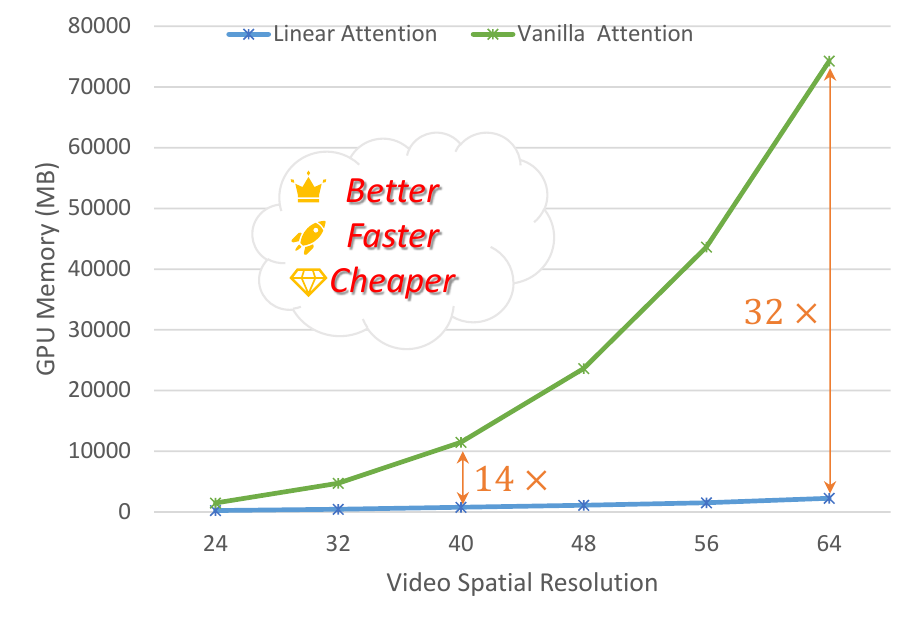}
  }

\caption{\textbf{The analysis of throughput and GPU memory consumption.} All input videos consist of 8 frames. Testing is performed on an NVIDIA A800 (80GB) GPU with a batch size of 1. Our MiraMo with linear attention is better, faster, and cheaper for video generation.}
\label{fig:throughput_gpu}
\end{figure}

\label{related_work}
\textbf{Text-to-video generation} aims to produce high-quality videos given text descriptions as conditional inputs. In recent years, diffusion models~\cite{ho2020denoising,song2021denoising,song2021score}, flow matching models~\cite{esser2024scaling,lipman2022flow}, and autoregressive models have made significant progress in text-to-image (T2I) generation. Existing T2I methods are capable of generating realistic images closely aligned with textual prompts \cite{ramesh2021zero,saharia2022photorealistic,yu2022scaling,ramesh2022hierarchical,chen2023pixart,tian2024visual,han2025infinity,ma2025training}. The following T2V generation approaches primarily involve augmenting T2I methods with additional temporal modules such as temporal convolutions or temporal attentions, or by using full attention mechanisms to establish temporal relationships between video frames~\cite{wang2023lavie,guo2024animatediff,ma2024latte,wu2023tune,blattmann2023align,ge2023preserve,zhou2022magicvideo,he2022latent,villegas2022phenaki,polyak2024movie,kong2024hunyuanvideo,deng2024autoregressive,jin2024pyramidal,tang2025spatial,yuan2025magictime,hu2025cascaded,ma2025efficient}. Due to the scarcity of available high-resolution clean video data, most of these methods rely on joint image-video training \cite{ho2022video} and typically build their models on pre-trained image models such as Stable Diffusion \cite{rombach2022high}, DALL$\cdot$E2 \cite{ramesh2022hierarchical}, proprietary internal models.
In terms of model architecture, current T2V techniques primarily focus on two designs: one is a cascaded structure \cite{ho2022imagen,ge2023preserve,singer2022make} inspired by \cite{ho2022cascaded} and the other is based on latent diffusion models \cite{blattmann2023align,zhang2023i2vgen,zhou2022magicvideo} extending the success of \cite{rombach2022high} to the video domain. In terms of the model purpose, these approaches can be classified into two major categories. Firstly, most methods aim at learning general motion representations, which typically rely on large and diverse datasets \cite{wang2023lavie,ma2024latte,blattmann2023align,blattmann2023stable}. Secondly, another stream of research focuses on personalized video generation. They focus on fine-tuning the pre-trained T2I or T2V models on narrow datasets customized for specific applications or domains \cite{guo2024animatediff,wu2023tune}. Our proposed MiraMo can be regarded as a controllable T2V model that can be conditioned on either texts or images.

\begin{figure*}[ht]
    \centering
    \includegraphics[width=1.0\linewidth]{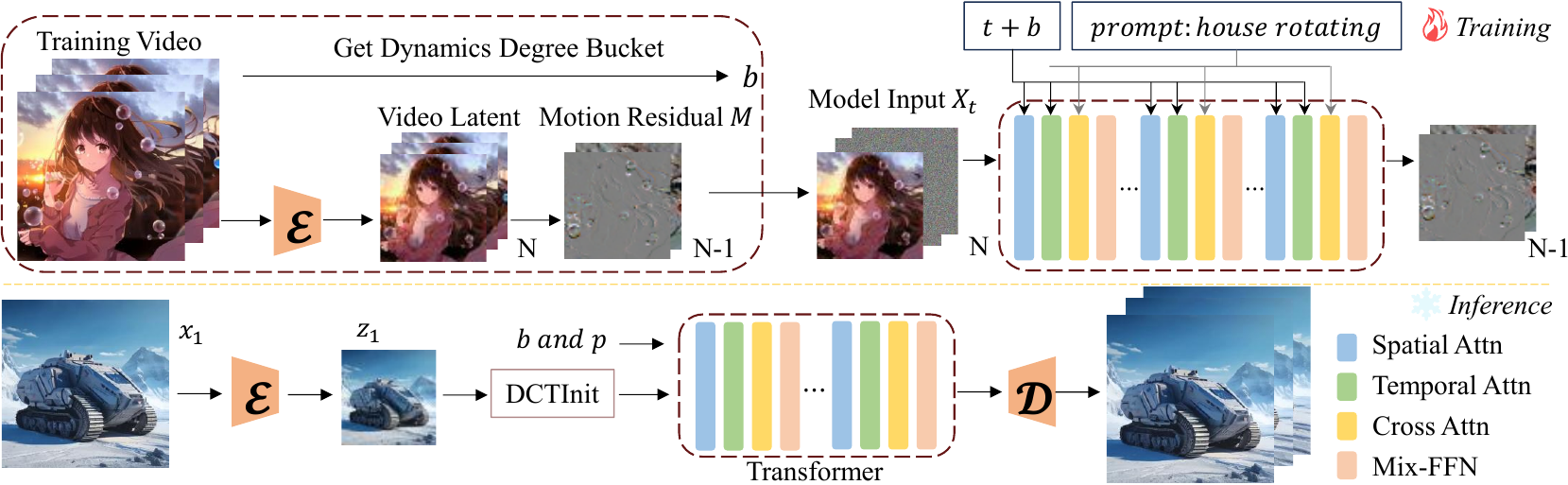}
    \caption{\textbf{Model pipeline overview.} Our model learns the distribution of motion residuals instead of predicting the frames directly. The details of the training procedure can be seen in the algorithm. \ref{alg:training}.}
    \label{fig_pipeline}
\end{figure*}

\textbf{Image animation} aims to maintain the identity of the static input image while crafting a coherent video and has gained attention in the fields of computer vision and computer graphics for decades. 
Early physics-based simulation methods focused on mimicking the motions of certain objects, such as human hair~\cite{xiao2023automatic}, talking heads \cite{geng2018warp,wang2020imaginator,wang2022latent,cui2025hallo3,zhou2025zero}, and human bodies \cite{bertiche2023blowing,blattmann2021understanding,siarohin2021motion,wang2023leo,wang2022latent,10645735,tu2025stableanimator,Chang_2025_CVPR}, etc., but their lack of generalization stemmed from separately modeling each object category \cite{chuang2005animating,xu2008animating,dorkenwald2021stochastic,prashnani2017phase,siarohin2021motion}. 
Subsequent GAN-based approaches overcome the manual segmentation, enabling the synthesis of more natural movements \cite{wang2022latent,wu2021f3a,chen2020animegan,pumarola2018ganimation}.
Recent methods mostly rely on foundational T2I or T2V pre-trained models, using RGB images as additional conditions to generate video frames from input images. Initially, VideoComposer \cite{wang2024videocomposer} and VideoCrafter1 \cite{chen2023videocrafter1} utilize the CLIP \cite{radford2021learning} embedding of the input static image as an additional condition for the T2V model to achieve image animation. Later works such as AnimateLCM~\cite{wang2024animatelcm} and DreamVideo~\cite{wang2023dreamvideo} found that the CLIP image encoder often misses fine details of the input image, leading to less faithful generation. To address this, they use lightweight networks to extract additional image embeddings as supplementary conditions for the base T2V model. Other mask-based methods, such as SEINE \cite{chen2023seine}, VDT \cite{lu2023vdt}, and Through-The-Mask~\cite{Yariv_2025_CVPR}, take different approaches to incorporating masking for image-to-video generation. SEINE and VDT apply random masking to input frames during training, enabling the prediction of future video frames from a single image. In contrast, Through-The-Mask uses masks as intermediate motion trajectories to capture both semantic object information and motion.
Plug-to-play methods like I2V-adapter \cite{guo2023i2v} and PIA \cite{zhang2023pia} utilize pre-trained LoRA~\cite{hu2022lora} weights to animate the input images. 
However, these diffusion-based methods cannot guarantee the appearance consistency and motion smoothness of the given images. 
By focusing on specific domains such as human body or avatar motion, methods like AnimateAnyone, MagicAnimate, etc.,~\cite{hu2023animate, xu2024magicanimate,tu2025stableanimator,Chang_2025_CVPR,cui2025hallo3} leverage additional motion cues such as pose sequences or audio to drive the movement of a given image and generate high-quality videos. But they hardly extend to other domains.
In contrast, our MiraMo does not restrict the domain of the input images, while suppressing abrupt motion changes via multiple tailored strategies.

\textbf{Efficient diffusion models} have tackled the challenges of large model sizes and long inference times. Recent work has explored architectural optimizations to eliminate redundancy in large models, enabling on-device generation within seconds. Both SnapFusion~\cite{li2023snapfusion} and MobileDiffusion~\cite{zhao2024mobilediffusion} propose an efficient U-Net by identifying redundant modules and conducting a comprehensive analysis of each component, respectively. BitsFusion~\cite{sui2024bitsfusion} then apply weight quantization to reduce parameters to 1.99 bits, achieving a 7.9× smaller model size while maintaining competitive generation quality. However, these models are generally constrained to low-resolution outputs, typically $512 \times 512$ pixels. To support efficient high-resolution generation, methods such as SANA~\cite{xie2024sana} and LinFusion~\cite{liu2024linfusion} employ linear attention to achieve 1K or larger resolution generation even on laptop GPUs. SnapGen~\cite{chen2025snapgen} generates 1K images on a mobile device in approximately 1.4 seconds by systematically examining network architecture design choices.  While these advancements have primarily focused on image generation, progress in video generation remains limited. Matten~\cite{gao2024matten} and LinGen~\cite{wang2025lingen} use Mamba blocks~\cite{gu2023mamba,dao2024transformers} to replace the vanilla attention modules in DiT blocks~\cite{peebles2023scalable}. The main distinction of MiraMo compared to the above methods is its focus on exploring linear attention for video generation.

%% file: sections/method.tex
\section{Methodology}
\label{methodology}
We start with a brief introduction of latent diffusion models, flow matching, video latent diffusion models, and image animation formulation. Following that, we present our designs for a self-constructed T2V linear Transformer model, motion residual learning, DCT-based inference noise refinement strategy, and dynamics degree control. The overall pipeline of our model is shown in Fig. \ref{fig_pipeline}.

\subsection{Preliminary and Problem Formulation}
\label{method:preliminary}
\textbf{Latent diffusion models} (LDMs) are efficient diffusion models that operate the diffusion process in the latent space of the pre-trained variational autoencoder (VAE) rather than the pixel space~\cite{song2021denoising,rombach2022high,ho2020denoising,kingma2013auto,kingma2019introduction}. An encoder $\mathcal{E}$ from the pre-trained VAE is firstly used in LDMs to project the input data sample $x \in p_{\rm data}$ into a lower-dimensional latent code $z = \mathcal{E}(x)$. Then, the data distribution is learned through two key processes: diffusion and denoising. The diffusion process gradually adds Gaussian noise into the latent code $z$ and the perturbed sample $z_{t} = {\sqrt{\overline{\alpha}_{t}}}z + \sqrt{1-{\overline{\alpha}_{t}}}\epsilon$, where $\epsilon\sim \mathcal{N}(0,1)$, following a $T$-stage Markov chain, is obtained. Here, $\overline{\alpha}_{t}$ and $t$ are the pre-defined noise scheduler and the diffusion timestep, respectively. The denoising process learns to inverse the diffusion process by predicting a less noisy sample $z_{t-1}$: $p_\theta(z_{t-1}|z_t)=\mathcal{N}(\mu_\theta(z_t),{\Sigma_\theta}(z_t))$ and make the variational lower bound of log-likelihood reduces to $\mathcal{L_\theta}=-\log{p(z_0|z_1)}+\sum_tD_{KL}((q(z_{t-1}|z_t,z_0)||p_\theta(z_{t-1}|z_t))$. In this context, $\mu_\theta$ means a denoising model and is trained with the objective,
\begin{equation}
\label{equ_l_simple}
\mathcal{L}_{\rm simple} = \mathbb{E}_{\mathbf{z}\sim p(z),\ \epsilon \sim \mathcal{N} (0,1),\ t}\left [ \left \| \epsilon - \epsilon_{\theta}(\mathbf{z}_t, t)\right \|^{2}_{2}\right].
\end{equation}

\textbf{Flow matching} follows the common concepts of the standard diffusion models (i.e., DDPMs). The diffusion formulation $z_t = tz_1 + (1- t)z_0$, where $z_1$ means Gaussian noise and $z_t$, can be regarded as a linear interpolation between $z_0$ and $z_1$~\cite{lipman2022flow,esser2024scaling}. Unlike DDPMs, which typically rely on noise prediction, the flow matching training objective is based on velocity prediction, using the ground truth velocity $v_t = \frac{dz_t}{dt} = z_1 - z_0$, and can be mathematically formulated as follows,
\begin{equation}
\mathcal{L} = \mathbb{E}_{\mathbf{z}\sim p(z),\ t}\left [ \left \| v_t - u_{\theta}(\mathbf{z}_t, t)\right \|^{2}_{2}\right],
\label{equ_flow_matching}
\end{equation}
where $u_\theta$ means a learnable model parameterized by $\theta$. 

\textbf{Video latent diffusion models} (VLDMs) extend the LDMs to a video counterpart by introducing the temporal motion module to build the temporal relationship between frames, combined with either DDPM loss or flow matching loss~\cite{wang2023lavie,ma2024latte,chen2023seine,wang2024videocomposer}. Based on the above design philosophy, MiraMo extends the SANA~\cite{xie2024sana} T2I model by incorporating a temporal module, leading to a T2V model built upon a linear Transformer. Given a video clip $\mathbf{V} \in \mathbb{R}^{N \times C \times H \times W}$, where $N, C, H, W$ denote the number of frames, the number of channel, height, and width respectively, we first project it into a latent space frame-by-frame to obtain the latent code. After that, the flow matching diffusion and denoising processes are performed in this latent space. Finally, the videos are generated through a decoder.

\textbf{Image animation formulation.} Given a static input image $x_1 \in \mathbb{R}^{C \times H \times W} $ and a textual prompt $p$, image animation aims at generating an $N$-frame video clip $\mathbf{V} = \{x_1, x_2, x_3, ..., x_N\}$, where $x_1$ is the first frame of the video and $x_i\in\mathbb{R}^{C \times H \times W}$.
The appearance of the subsequent frames $\{x_2, x_3, ..., x_N\}$ should be closely aligned with $x_1$, while the content and motion of the video adhere to the textual description provided in $p$. 
We decompose this problem into three sub-problems: learning the motion residuals, suppressing the abrupt motion changes, and controlling the degree of motion dynamics.

\begin{figure}[ht]  
\centering
\subfloat[SANA block]
  {
    \includegraphics[scale=1.0]{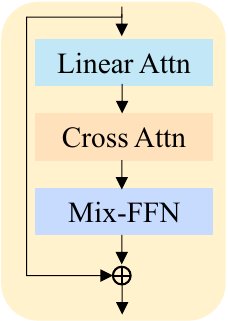}
  }
\subfloat[Mix-FFN]
  {
    \includegraphics[scale=1.0]{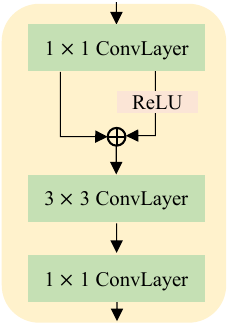}
  }

\caption{The structure of the SANA block and Mix-FFN.}
\label{fig:sana_block}
\end{figure}

\subsection{The Self-constructed T2V Linear Transformer Model}
\label{the_self-constructed_t2v_linear_transformer_model}
Following the common practice of extending a base T2I model to a T2V model, we explore two different ways to adapt the SANA~\cite{xie2024sana} T2I model for video generation.

\textbf{Add temporal convolutional layer.} Fig.~\ref{fig:sana_block} shows the core block of the SANA model that contains a linear self-attention, a cross attention, and a Mix-FFN. The SANA block is similar to the DiT block, with the key differences being the replacement of vanilla self-attention with more efficient linear self-attention, and the substitution of the standard FFN with a MiX-FFN implemented using efficient 2D convolutional layers. SANA enables generation without relying on any explicit positional embeddings, while still allowing the model to perceive the spatial positions of tokens. This is similar to what has been observed in the Stable Diffusion family of models, which also operates without positional embeddings. The underlying reason for this phenomenon is that the 2D convolutional layers in the model inherently capture spatial positional information. Inspired by this phenomenon, we add an additional temporal convolutional layer after the MiX-FFN to enable the model to capture temporal relationships in videos. However, we find that this approach quickly degrade the original image generation capability of SANA. Moreover, training only the newly added parameters fails to converge. We believe the failure may be attributed to the fixed local receptive field of temporal convolution operations, which imposes a strong local inductive bias. Forcibly mixing adjacent frames through such operations can disrupt the original spatial relationships.

\textbf{Add temporal attention layer.} To overcome this, we turn to a more flexible attention mechanism. Following prior works~\cite{kong2024hunyuanvideo,wan2025wan,lin2024open}, we incorporate Rotary Positional Embeddings (RoPE) to encode temporal positions. Next, we introduce a form of linear attention that is different from SANA and compatible with RoPE.

Recall that the scaled dot-product attention mechanism can be written as,
\begin{equation}
\mathcal{A}(Q, K, V) = \sigma(QK^T)V 
\label{equation_scaled_dot_attention}
\end{equation}
where $\mathcal{A}$ and $\sigma$ are attention operations and the sigmoid function, respectively. For simplicity, we omit the explicit scaling factor in the attention formulation. Equivalently, the attention for a single query vector $q_i$ can be written as,
\begin{equation}
\mathcal{A}(Q, K, V)_i = \frac{\sum_{j=1}^n e^{q_i k_j^T} v_j}{\sum_{j=1}^{n} e^{q_i k_j^T}},
\label{equation_scaled_dot_attention_vector_form}
\end{equation}
where $q_i, k_j, v_j \in \mathbb{R}^{1 \times d}$ are the row vectors. 
More generally, the dot-product similarity can be replaced by any non-negative similarity function~\cite{wang2018non} $sim(\cdot, \cdot)$,
\begin{equation}
\mathcal{A}(Q, K, V)_i = \frac{\sum_{j=1}^n sim(q_i, k_j) v_j}{\sum_{j=1}^{n} sim(q_i, k_j)}.
\label{equation_scaled_dot_attention_vector_form_sim}
\end{equation} 

In SANA, a ReLU activation is applied to ensure non-negativity: $sim(q_i, k_j) = \phi(q_i)\phi(k_j)^T$ with $\phi = \text{ReLU}$. This results in the linear attention form,
\begin{equation}
\mathcal{A}(Q, K, V)_i = \frac{\sum_{j=1}^n \phi(q_i)\phi(k_j)^T v_j}{\sum_{j=1}^{n} \phi(q_i)\phi(k_j)^T}.
\label{equation_sana_attention_vector_form}
\end{equation} 

However, directly applying the rotation matrix ($\mathcal{R}$) of RoPE to this form (Eq.~\ref{equation_sana_attention_vector_form_rope}) may introduce negative inner products, which violates the non-negativity required for probabilistic interpretation and may lead to instability, as the denominator could approach zero,
\begin{equation}
\mathcal{A}(Q, K, V)_i = \frac{\sum_{j=1}^n \left [\mathcal{R}_i\phi(q_i) \right ] \left [\mathcal{R_j}\phi(k_j) \right ]^T v_j}{\sum_{j=1}^{n} \left [\mathcal{R}_i\phi(q_i) \right] \left [\mathcal{R_j}\phi(k_j) \right ]^T}.
\label{equation_sana_attention_vector_form_rope}
\end{equation} 

To resolve this, we adopt a cosine-similarity-based attention inspired by the success of RoFormer and its variants~\cite{rope,sujianlin-linear} in the natural language process (NLP), and we design,
\begin{equation}
sim(q_i,k_j) = 1 + (\frac{q_i}{\lVert q_i \rVert}) (\frac{k_j}{\lVert k_j \rVert})^T.
\label{equation_cosine_similarity}
\end{equation}

The RoPE operation can be directly applied to Equ.~\ref{equation_cosine_similarity} as this form of $sim(q_i,k_j)$ does not rely on the non-negativity of the values, and RoPE does not change the vector magnitudes. Finlay, the temporal linear attention formulation can be written as follows mathematically,
\begin{equation}
\mathcal{A}(Q, K, V)_i = \frac{\sum_{j=1}^nv_j + \frac{q_i}{\lVert q_i \rVert}\left [\sum_{j=1}^n(\frac{k_j}{\rVert K_j \rVert})^Tv_j \right ]}{n + \frac{q_i}{\rVert q_i \rVert}\sum_{j=1}^n(\frac{k_j}{\rVert K_j \rVert})^T}.
\label{equation_final_linear_attention}
\end{equation}

Similar to SANA, the shared terms $\sum_{j=1}^n(\frac{k_j}{\rVert K_j \rVert})^Tv_j$ and $\sum_{j=1}^n(\frac{k_j}{\rVert K_j \rVert})^T$ can be computed only once and reused for each query, leading to savings in computation and memory.

We insert a temporal attention layer between the linear attention and cross attention of the SANA block, and explore two different approaches to temporal modeling using Eq.\ref{equation_final_linear_attention}. The first approach, inspired by Latte\cite{ma2024latte}, LaVie~\cite{wang2023lavie}, Dynamicrafter~\cite{xing2023dynamicrafter}, and others, applies temporal attention only across the frame dimension of videos. The second approach, following works like CogVideoX~\cite{yang2024cogvideox}, LTX-Video~\cite{hacohen2024ltx}, and HunyuanVideo~\cite{kong2024hunyuanvideo}, performs temporal attention over all video tokens in a fully 3D manner with 3D RoPE. Both temporal modeling strategies are implemented under a unified image-video joint training framework. We find that the first approach outperforms the second in terms of temporal consistency and others.

\subsection{Motion Residual Learning} 
\label{method_image_animation_formulation}
Once the T2V model is trained, we discuss here how to achieve image animation. Instead of directly predicting the frames as in previous methods, such as PIA~\cite{zhang2023pia}, ConsistI2V~\cite{ren2024consisti2v}, I2V-Adapter~\cite{guo2023i2v}, and DynamiCrafter~\cite{xing2023dynamicrafter}, we propose to learn the latent-domain differences ({\em motion residuals}) across the frames. 
Learning the residuals allows the model to focus on the changes among frames, instead of repeated or redundant information that appears in all frames. 
Hence, the model can better capture the dynamics feature in the video, and lower the chance of predicting discontinuous or abrupt motion.  

Specifically, we sample an $N$-frame video clip $\mathbf{V} = \{x_1, x_2, x_3, \ldots, x_N\}$, from the training dataset and set the first frame $x_1$ as the image to be animated. As described in Sec.~\ref{method:preliminary}, we first utilize a pretrained VAE~\cite{xie2024sana} to compress the video clip $\mathbf{V}$ into a low-dimensional latent space to obtain the latent code $\mathbf{Z} = \{z_1, z_2, z_3, ..., z_N\}$, where $z_i\in\mathbb{R}^{c\times h\times w}$. Here, $c$, $h$, and $w$ denote the channel, height, and width of the frame in latent space, respectively.

Next, we compute the motion residuals $\mathbf{M}=\{z_2-z_1,z_3-z_1,\ldots,z_N-z_1\}$, by subtracting the first frame from all subsequent frames. To guide the model to predict motion residuals, at each diffusion timestep $t$, we first add the noised motion residuals $\mathbf{M}_t$ to the features of the input image $z_1$ to obtain $\mathbf{M}_t^{'}$. Then, we concatenate $z_1$ and $\mathbf{M}_t^{'}$ to form $N$ frames, which are used as the input $\mathbf{X}_t$ to the model. We select the last $N - 1$ frames output by the model as the denoised $\mathbf{M}_{t-1}$. The detailed training procedure of our model is summarized in Algorithm. \ref{alg:training} and we assume that batch size is set to 1.

\begin{algorithm}[!h]
\caption{The training procedure.}
\label{alg:training}
\begin{algorithmic}[1]
\Repeat
    \State Sample video $\mathbf{V} =\{x_1,\ldots,x_N\}$ from training set, where $x_i\in \mathbb{R}^{C \times H \times W}$ is a frame
    \State Compute dynamics degree bucket $b$ from $\mathbf{V}$ (Eq.\ \ref{eq:b})
    \State Compute $\mathbf{Z} = \{z_1, \ldots, z_n\}$, where $z_i=\mathcal{E}(x_i)\in\mathbb{R}^{c \times h \times w}$ for each $x_i\in\mathbf{V}$
    \State Compute $\mathbf{M} = \{z_2-z_1, z_3-z_1, \ldots, z_N-z_1  \}$
    \State Sample $t \sim \text{Uniform}(1,\ldots,T)$
    \State Sample $\epsilon \sim \mathcal{N}(0, I)$
    \State Get noised $\mathbf{M}_t$ via the diffusion process
    \State Get input $\mathbf{X_t}$ = \textit{torch.cat}([$z_1$, $\mathbf{M}_t$+$z_1$])
    \State Take gradient descent step on Eq.\ \ref{equ_flow_matching}
\Until{Converged}
\end{algorithmic}
\end{algorithm}

\subsection{DCT-based Noise Refinement}
Recent literature~\cite{lin2024common,si2024freeu} have identified inconsistencies between the noises, used during the training and inference phases of DDPMs, which may negatively impact the sampling outcomes. 
These inconsistencies stem from the leakage of low-frequency information. 
Specifically, the noise during training contains low-frequency data from the training samples, while the noise used during testing lacks such information.
This issue still exists in the flow matching, as we typically follow the logit-normal timestep schedule used in Stable Diffusion 3~\cite{esser2024scaling}, which results in a very low probability of the SNR reaching zero.
To address this issue, existing I2V methods~\cite{ren2024consisti2v,wu2023freeinit} employ the Fast Fourier Transform (FFT) to inject low-frequency components of input images into the original noise, and significantly enhance the generation results.

\begin{figure}[ht]
    \centering
    \includegraphics[width=0.9\linewidth]{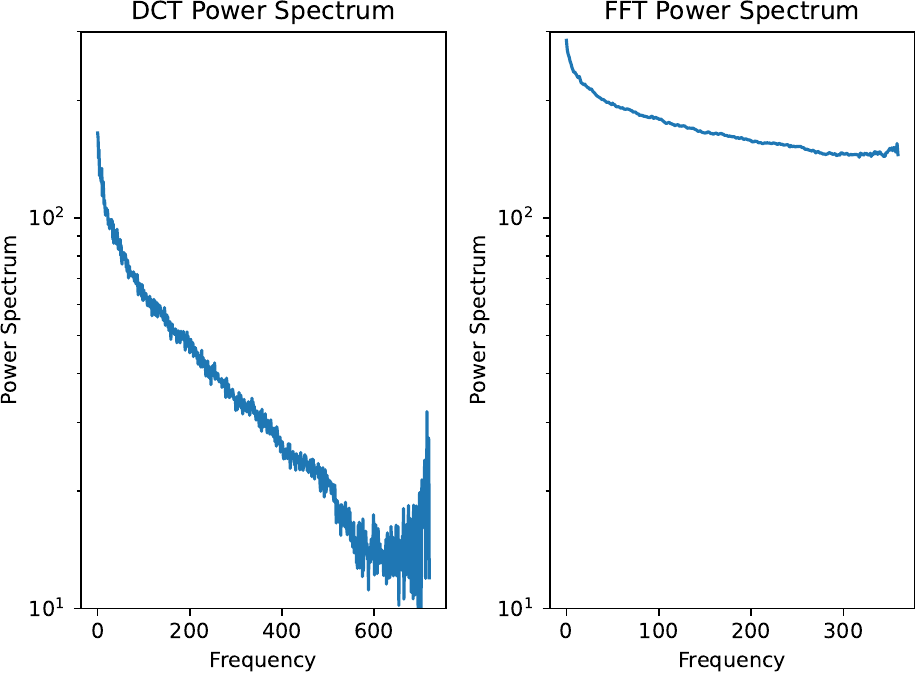}
    \caption{The power spectrums of DCT and FFT.}
    \label{fig_power_spectrum}
\end{figure}

However, we observe that this approach may still lead to problems such as inconsistent video color, as shown in Fig.~\ref{fig_fft_vs_dct_e_dctinit}(a) (please visit the project page to visualize such inconsistency). We attribute this to two main factors: (1) spectral leakage due to edge effects since FFT assumes the input signal is periodic, and the finite length of input images results in these edge effects; and (2) poor concentration of low-frequency energy. 
The complex exponential basis functions used in the transformation disperse the energy of low-frequency components as shown in Fig. \ref{fig_power_spectrum}, leading to insufficient low-frequency information in the injected noise. 
To overcome these challenges, we propose the use of the Discrete Cosine Transform (DCT), termed DCTInit. The symmetric periodic extension and energy-packing nature of low-frequency components of DCT effectively address the above two issues.

Mathematically, given the latent code $z_1$ of the input static image $x_1$ and the inference noise $\epsilon$, we first add $\tau$-step inference noise to $z_1$, leading to $z^{\tau}_1$. We then extract the low-frequency DCT coefficients $\mathcal{D}_{z^{\tau}_1}^L$ of $z^{\tau}_1$ and mix it with the high-frequency DCT coefficients $\mathcal{D}_{\epsilon}^H$ of $\epsilon$,
\begin{equation}
    \mathcal{D}_{z^{\tau}_1}^L = \mathcal{DCT}_{\rm 3D}(z^{\tau}_1) \odot \mathcal{H},
\end{equation}
\begin{equation}
    \mathcal{D}_{\epsilon}^H = \mathcal{DCT}_{\rm 3D}(\epsilon) \odot (1-\mathcal{H}),
\end{equation}
\begin{equation}
    \epsilon' = \mathcal{IDCT}(\mathcal{D}_{z^{\tau}_1}^L + \mathcal{D}_{\epsilon}^H).
\end{equation}
Here, $\mathcal{DCT}_{3D}$ is the DCT operated on both spatial and temporal dimensions, $\mathcal{IDCT}$ is the inverse DCT operation, and $\mathcal{H}$ is the low pass filter. The refinement noise $\epsilon'$, which contains the low-frequency information of $z_1$, is then used for denoising. As shown in Fig.~\ref{fig_fft_vs_dct_e_dctinit}(b), DCTInit can improve the temporal consistency and mitigate sudden motion changes in generated videos.

\subsection{Dynamics Degree Control}
Although the previous two designs can alleviate the abrupt appearance inconsistency and discontinuous motion, they may also tend to reduce the motion dynamics that could be generated. 
To allow explicit and controlled dynamics of the generated video content, we introduce a simple and effective strategy, inspired by~\cite{dai2023animateanything}, that uses the average multi-scale structural similarity (MS-SSIM) $s$ between frames as a means to finely control the degree of motion dynamics:
\begin{equation}
    s(\mathbf{V}) = \frac{1}{N-1}\sum_{i=2}^N {\rm MS\_SSIM}(x_i, x_{i-1}).\label{eq:b}
\end{equation}
Here, the dynamics degree $s$ measures the differences between frames in the pixel space; $N$ is the total number of frames, in our case, $N$=16. We find that the dynamics degree calculated from video clips sampled with a fixed frame interval exhibits a significant long-tail distribution. To alleviate this skewness, we randomly select a frame interval between frames 3 and 10 to sample video clips. After obtaining the dynamics degree $s$, we uniformly project $s$ into the dynamics degree bucket $b$ (ranging from 0 to 19). Similar to the timestep $t$, we project the bucket $b$ onto the positional embedding, then add it to the timestep embedding, and finally incorporate it into each frame via the single AdaIN layer to ensure that the dynamics degree is applied uniformly across all frames.

Finally, combining the motion residuals learning, the dynamic degree control, and Eq.~\ref{equ_flow_matching}, the final learning objective is formulated as:
\begin{equation}
\mathcal{L} = \mathbb{E}_{\mathbf{z}\sim p(z),\ t}\left [ \left \| v_t - u_{\theta}(\mathbf{X}_t, p, b, t)\right \|^{2}_{2}\right],
\label{equ_miramo_objective}
\end{equation}

\subsection{Synthetic Video Data for the post-training}
\label{sec_synthetic_video_data_for_quality_fine-tuning}
Previous studies have highlighted the importance of the post-training stage, which involves supervised fine-tuning (SFT) on a small but high-quality dataset following large-scale pretraining. This process can significantly improve video generation performance, including frame-level visual quality and instruction-following capabilities. However, collecting such high-quality datasets remains a major challenge.
Meanwhile, current large-scale video generation models are already capable of producing reasonably high-quality videos. Based on this observation, we use two T2V models, Wan2.1~\cite{wan2025wan} and HunyuanVideo~\cite{kong2024hunyuanvideo}, to generate approximately 150,000 videos as a substitute for the desired small but high-quality video dataset. We then perform the supervised fine-tuning on both the generated dataset and a real high-quality dataset of similar scale. The results show that the model fine-tuned on the generated dataset outperforms the one trained on the real dataset. Please refer to Sec.~\ref{sec_the_benefit_of_synthetic_video_data} for more details.

%% file: sections/experiment.tex
\section{Experiments}

\begin{figure*}[!t]
\centering
\includegraphics[width=1.0\linewidth]{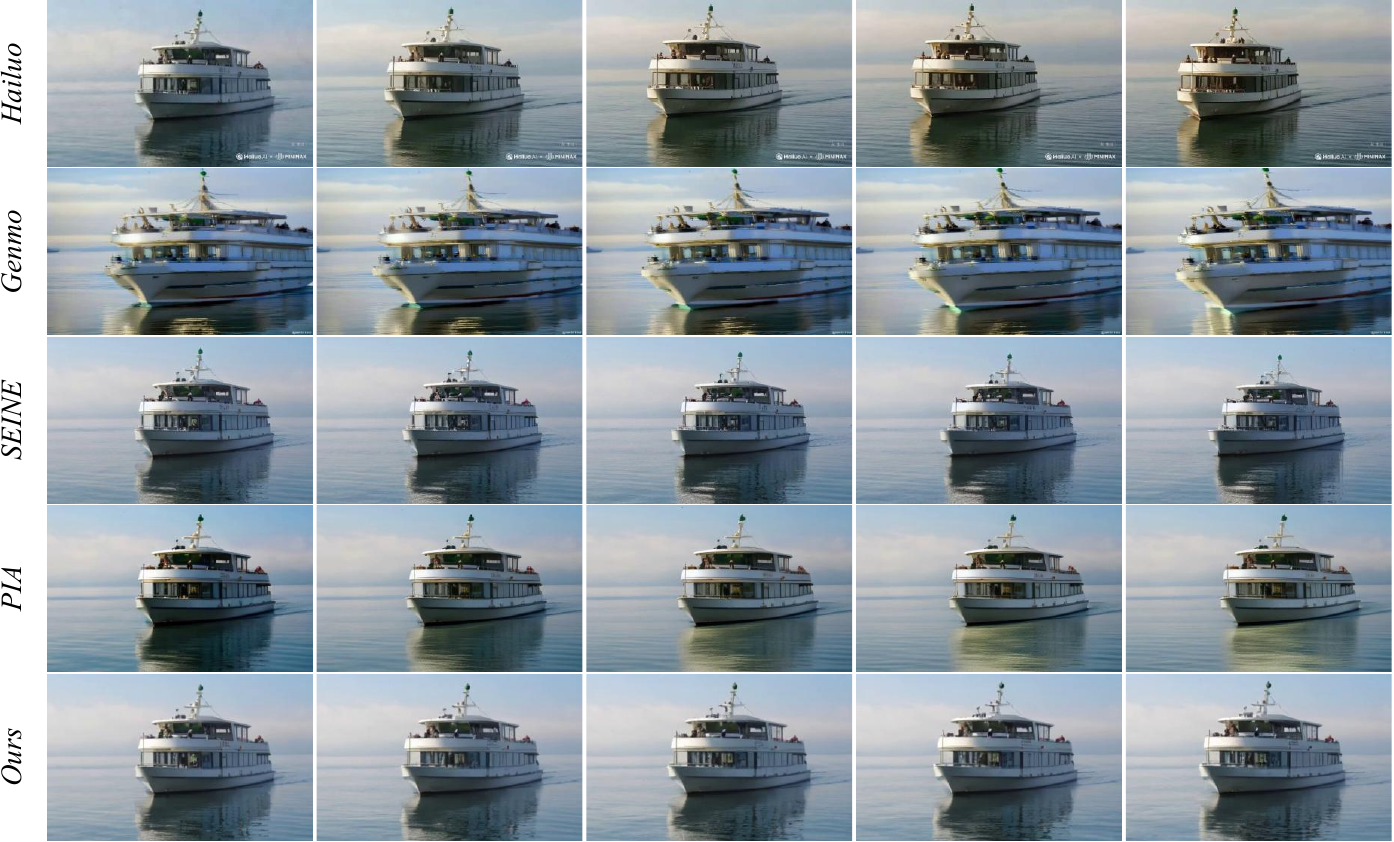}
\caption{\textbf{Qualitative visual comparisons}. We compare Cinemo with both closed-source commercial tools and open-source research works. ``{\tt\small a ship sailing on the water}'' is the user prompt. Our videos are cropped for better comparison. \textit{Please visit the project page to visualize the complete animation comparisons.}}
\label{fig:qualitative_comparison}
\end{figure*}

\label{experiments}
We first outline the experimental setup, covering datasets, baselines, evaluation metrics, and implementation details. Following that, we compare the experimental results with state-of-the-art methods. Finally, we present the analysis of our method.

\subsection{Experimental Setup}
\label{experiments_experimental_setup}
\textbf{Datasets and implementation details.} We first train our self-constructed T2V linear transformer model on a subset of the Vimeo25M video dataset~\cite{wang2023lavie}, consisting of approximately 2,000,000 videos.
For each video, we randomly sample 16 frames (i.e., $N=16$) at a spatial resolution of $512 \times 512$ pixels, with a frame interval ranging from 3 to 10 frames. This sampling strategy ensures a consistent distribution of motion dynamics as conditional input to our model. We use HunyuanVideo~\cite{kong2024hunyuanvideo} and Wan2.1~\cite{wan2025wan} to generate approximately 150,000 videos, forming a small yet high-quality video dataset for the post-training stage. As for image animation training, the first frame is regarded as the input static image, and our model is trained to denoise the motion residuals of the subsequent 15 frames. During the inference, we use the Flow-DPM-Solver \cite{xie2024sana} with 25 steps and apply classifier-free guidance with a guidance scale of 5.5 to animate images in our experiments. The model architecture of MiraMo is identical to SANA~\cite{wang2023lavie}, with the
addition of a temporal attention layer for capturing temporal relationships in videos and a linear layer for projecting dynamics degree buckets $b$ into embeddings. The entire model is optimized using the Adam optimizer on 8 NVIDIA A800 (80GB) GPUs, using a total batch size of 56.
\textbf{}

\begin{table*}[!t]
\centering
\resizebox{\linewidth}{!}{
\begin{tabular}{cccccccccc}
\hline
Methods                                           & \begin{tabular}[c]{@{}c@{}}Subject\\ Consistency\end{tabular} & \begin{tabular}[c]{@{}c@{}}Background\\ Consistency\end{tabular} & \begin{tabular}[c]{@{}c@{}}Temporal\\ Flickering\end{tabular} & \begin{tabular}[c]{@{}c@{}}Motion\\ Smoothness\end{tabular}    & \begin{tabular}[c]{@{}c@{}}Dynamic \\ Degree\end{tabular} & \begin{tabular}[c]{@{}c@{}}Aesthetic\\ Quality\end{tabular} & \begin{tabular}[c]{@{}c@{}}Imaging\\ Quality\end{tabular} & \begin{tabular}[c]{@{}c@{}}Object\\ Class\end{tabular} & Params        \\ \hline
LaVie~\cite{wang2023lavie} & 91.41\% & \underline{97.47\%} & 98.30\% & 96.38\% & 49.72\% & 54.94\% & {61.90\%} & \textbf{91.82\%} & 0.91B \\
ModelScope~\cite{VideoFusion} & 89.87\%  & 95.29\% & 98.28\% & 95.79\% & {66.39\%} & 52.06\% & 58.57\% & 82.25\% & 1.41B \\
VideoCrafter2~\cite{chen2024videocrafter2} & \underline{95.10\%} & \textbf{98.04\%} & \underline{98.93\%} & 95.67\% & 55.00\% & \underline{62.67\%} & \underline{65.46\%} & 78.18\% & 1.85B\\
CogVideo \cite{hong2022cogvideo} & 92.19\% & 95.42\% & 97.64\% & {96.47\%} & 42.22\% & 38.18\% & 41.03\%  & 73.40\% & 9.40B \\
HiGen \cite{higen} & 90.07\% & 93.99\% & 93.24\% & \underline{96.69\%} & \textbf{99.17\%} & {57.30\%} & 63.92\% & 86.06\% & 4.60B \\
OpenSoraPlan~\cite{lin2024open} & \textbf{97.27\%} & 96.24\% & \textbf{99.12\%} & \textbf{99.17\%} & 35.28\% & 59.10\% & \textbf{65.73\%} & 61.53\% & 2.77B \\
Latte~\cite{ma2024latte} & 88.88\% & 95.40\% & 98.89\%  & 94.63\%  & \underline{68.89\%} & 61.59\%  & 61.92\% & 86.53\% & 1.05B \\ 
Ours & 94.35\% & 96.81\% & 96.06\% & 96.56\% & 66.39\% & \textbf{64.24\%} & 62.37\% & \underline{87.74\%} & 0.74B
\\ \hline \hline

Methods  & \begin{tabular}[c]{@{}c@{}}Multiple\\ Object\end{tabular}     & \begin{tabular}[c]{@{}c@{}}Human\\ Action\end{tabular}           & Color                                                         & \begin{tabular}[c]{@{}c@{}}Spatial\\ Relationship\end{tabular} & Scene                                                     & \begin{tabular}[c]{@{}c@{}}Appearance\\ Style\end{tabular}  & \begin{tabular}[c]{@{}c@{}}Temporal\\ Style\end{tabular}  & \begin{tabular}[c]{@{}c@{}}Overall\\ Consistency\end{tabular} & \begin{tabular}[c]{@{}c@{}}Average\\ Score\end{tabular} \\ \hline
LaVie~\cite{wang2023lavie} & 33.32\% & \textbf{96.80\%} & 86.39\% & 34.09\% & \textbf{52.69\%} & 23.56\% & \textbf{25.93\%} & 26.41\% & 63.82\% \\
ModelScope~\cite{VideoFusion} & 38.98\% & \underline{92.40\%} & 81.72\% & 33.68\% & 39.26\% & 23.39\% & 25.37\% & 25.67\% & 62.44\% \\
VideoCrafter2~\cite{chen2024videocrafter2} & \underline{45.66\%} & 91.60\% & \textbf{93.32\%} & \textbf{58.86\%} & 43.75\% & \underline{24.41\%} & \underline{25.54\%} & 26.76\% & \underline{66.18\%} \\
CogVideo \cite{hong2022cogvideo} & 18.11\%  & 78.20\%  & 79.57\% & 18.24\%  & \textbf{28.24\%}  & 22.01\% & 7.80\% & 7.70\% & 52.28\% \\
HiGen~\cite{higen} & 22.39\% & 86.20\% & 86.22\% & 22.43\% & 44.88\% & \textbf{24.54\%} & 25.14\% & 27.14\% & 63.71\% \\
OpenSoraPlan~\cite{lin2024open} & 24.91\% & 58.20\% & \underline{90.90\%} & \underline{49.64\%} & 17.28\% & 20.04\% & 19.14\% & 20.39\% & 57.12\% \\
Latte~\cite{ma2024latte} & \underline{34.53\%} & {90.00\%} & {85.31\%} & 41.53\% & 36.26\% & 23.74\% & 24.76\% & \underline{27.33\%} & 63.76\% \\
Ours & \textbf{54.01\%} & 90.80\% & 89.73\% & 40.18\% & \underline{52.01\%} & 23.83\% & 25.44\% & 28.24\% & \textbf{66.80\%} \\ \hline
\end{tabular}
}
\caption{\textbf{Vbench T2V evaluation results per dimension for different methods.} A higher score demonstrates better model performance for a certain dimension except for \textit{Params}. We use \textbf{bold} and \underline{underline} to mark the best and second model performances, respectively.}
\label{tab_t2v_vbench}
\end{table*}

\textbf{Baselines and evaluation metrics.} 
For the T2V quantitative evaluation, considering our computational resources and dataset size, we select the following methods as competitors: LaVie~\cite{wang2023lavie}, ModelScope~\cite{VideoFusion}, VideoCrafter2~\cite{chen2024videocrafter2}, CogVideo~\cite{hong2022cogvideo}, HiGen~\cite{higen}, OpenSoraPlan~\cite{lin2024open}, and Latte~\cite{ma2024latte}. The evaluation is primarily conducted using the VBench~\cite{huang2024vbench} T2V benchmark. 

As for the I2V comparison, we compare with recent state-of-the-art I2V methods, including SVD \cite{blattmann2023stable}, I2VGen-XL \cite{zhang2023i2vgen}, DynamiCrafter \cite{xing2023dynamicrafter}, SEINE \cite{chen2023seine}, ConsistI2V \cite{ren2024consisti2v}, PIA \cite{zhang2023pia} and Follow-Your-Click \cite{ma2024follow}. 
Additionally, we compare our model against the commercial tools, Hailuo by MiniMax and Genmo. 
Following recent works \cite{ren2024consisti2v,dai2023animateanything}, we evaluate our model on two public datasets MSR-VTT \cite{xu2016msr} and UCF-101 \cite{soomro2012dataset}. 
We utilize Fr{\'e}chet Video Distance (FVD) \cite{unterthiner2019fvd} and Inception Score IS \cite{saito2017temporal} for assessing the video quality, Fr{\'e}chet Inception Distance (FID) \cite{parmar2021buggy} for evaluating the frame quality, and the CLIP similarity (CLIPSIM) \cite{wu2021godiva} for measuring the video-text alignment. 
We evaluate FVD, FID, and IS on UCF-101 using 2,048 random videos, and FVD and CLIPSIM on the MSR-VTT test split, which consists of 2,990 videos. 
Our primary focus is to assess the animation capability of our model. 
We select a random frame from a video snippet sourced from our evaluation dataset. This frame, combined with a textual prompt, is used as the input to generate animated videos in a zero-shot manner. 
To provide a more comprehensive automatic evaluation, we further compare our method against competitors using the metrics from the VBench benchmark~\cite{huang2024vbench} tailored for the I2V task.
Considering the inherently subjective nature of video generation assessment, we also conduct a user study following LaVie~\cite{wang2023lavie}. In this study, 10 raters individually evaluate each video based on five pre-defined criteria: motion smoothness, motion reasonableness, subject consistency, background consistency, and dynamic degree. They can grade ``good'', ``normal'', or ``bad'' for each criterion. All human evaluations are conducted without time restrictions. By combining metrics on UCF-101 and MSR-VTT, the VBench metrics and user study, we can better evaluate appearance consistency and motion smoothness.

\subsection{Comparisons with State-of-the-art Methods} 
\label{experiments_comparisons_with_state_of_the_art} 

\begin{table}[!t]
\centering
\resizebox{\linewidth}{!}{
\begin{tabular}{lccccc}
\toprule
\multirow{2}{*}{Method} & \multicolumn{3}{c}{UCF-101} & \multicolumn{2}{c}{MSR-VTT} \\ \cmidrule(lr){2-4} \cmidrule(lr){5-6} 
& FVD $\downarrow$ & IS $\uparrow$ & FID $\downarrow$ & FVD $\downarrow$ & CLIPSIM $\uparrow$ \\ \midrule

I2VGen-XL~\cite{zhang2023i2vgen} & 597.42 & 18.20 & 42.39 & 270.78 & 0.2541 \\
DynamiCrafter~\cite{xing2023dynamicrafter} & 404.50 & 41.97 & 32.35 & 219.31 & 0.2659 \\
SEINE~\cite{chen2023seine} & 306.49 & 54.02 & 26.00 & 152.63 & {0.2774} \\
ConsistI2V~\cite{ren2024consisti2v} & {177.66} & {56.22} & {15.74}  & {104.58} & 0.2674 \\ 
Follow-Your-Click~\cite{ma2024follow} & - & - & - & 271.74 & - \\
Cinemo~\cite{ma2025consistent}  & \textbf{168.16} &  58.71 & 13.17 & \textbf{93.51} & 0.2858 \\
Ours & 201.90 & \textbf{59.64} & \textbf{12.31} & 93.91 & \textbf{0.2903} \\ \hline
\end{tabular}
}
\caption{\textbf{Image animation quantitative comparisons.} $\downarrow$ means the lower the better. $\uparrow$ means the higher the better.} 
\label{table_ucf_msrvtt}
\end{table}

\begin{table}[!t]
\centering
\resizebox{\linewidth}{!}{
\begin{tabular}{cccccc}
\hline
Methods     & \begin{tabular}[c]{@{}c@{}}Imaging \\ Quality\end{tabular} & \begin{tabular}[c]{@{}c@{}}Aesthetic \\ Quality\end{tabular} & \begin{tabular}[c]{@{}c@{}} Camera \\ Motion\end{tabular} & \begin{tabular}[c]{@{}c@{}}Motion \\ Smoothness\end{tabular} & \begin{tabular}[c]{@{}c@{}}Dynamic \\ Degree\end{tabular} \\ \hline
DynamiCrafter~\cite{xing2023dynamicrafter} & 62.28 & 58.71 & 20.92 & 97.83 & \underline{40.57} \\
SVD~\cite{blattmann2023stable} & 70.23 & 60.23 & - & \underline{98.12} & \textbf{43.17}
\\
ConsistI2V~\cite{ren2024consisti2v} & 66.92 & 59.00
 & \textbf{33.92} & 97.38 & 18.62 \\
SEINE~\cite{chen2023seine} & 70.97 & 58.42
 & 23.67 & 96.68 & 27.07 \\
Cinemo~\cite{ma2025consistent} & \textbf{71.30} & \underline{61.85}
 & \underline{30.68} & \textbf{98.21} & 30.68 \\ 
Ours & \underline{70.98} & \textbf{65.44} & 21.57 & 96.93 & 35.12 \\ \hline
\end{tabular}
}
\caption{\textbf{The VBench I2V quantitative comparisons.} Higher values for these metrics indicate better performance (in \%), with the \textbf{best} and \underline{second-best} values highlighted accordingly.}
\label{table_vbench_comparison}
\end{table}

\textbf{Image animation quantitative comparison.} From Tab.~\ref{table_ucf_msrvtt}, our model achieves the highest IS and FID scores on the UCF-101 dataset and the highest CLIPSIM score on the MSR-VTT dataset, while attaining comparable FVD values on both UCF-101 and MSR-VTT. Furthermore, as shown in Tab.~\ref{table_vbench_comparison}, our method excels in image quality and aesthetic quality while also delivering competitive results in motion smoothness. Although the dynamic degree of our method is lower compared to DynamiCrafter and SVD, it still outperforms in human evaluations, as indicated in Tab.~\ref{tab_huamn_eval_metrics}, where it receives the highest proportion of ``good'' votes. Additionally, the dynamic degree can be controlled, as discussed in Sec.~\ref{sec_discussions_and_further_applications}. Notably, in this user study, our method achieves the highest ratings in motion smoothness, motion reasonableness, subject consistency, and background consistency, while maintaining the lowest ``bad'' votes across all metrics.

\begin{table*}[!t]
\centering
\begin{tabular}{cccccccccccccc}
\hline 
& \multicolumn{3}{c}{SVD~\cite{blattmann2023stable}} && \multicolumn{3}{c}{SEINE~\cite{chen2023seine}} && \multicolumn{3}{c}{\textbf{Ours}} \\
Metrics & Bad & Normal & Good && Bad & Normal & Good && Bad & Normal & Good \\
\cmidrule{2-4}\cmidrule{6-8}\cmidrule{10-12}
Motion Smoothness & 0.33 & 0.42 & 0.25 && 0.19 & 0.43 & 0.38 && 0.18 & 0.40 & \textbf{0.42} \\
Motion Reasonableness & 0.51 & 0.28 & 0.21 && 0.37 & 0.26 & 0.37 && 0.22 & 0.31 & \textbf{0.47} \\
Subject Consistency & 0.21 & 0.37 & 0.42 && 0.34 & 0.24 & 0.42 && 0.21 & 0.29 & \textbf{0.50} \\
Background Consistency & 0.16 & 0.31 & 0.53 && 0.23 & 0.24 & 0.53 && 0.13 & 0.31 & \textbf{0.56} \\
Dynamic Degree &0.51 & 0.31 & 0.18 && 0.18 & 0.42 & 0.40 && 0.11 & 0.46 & \textbf{0.43} \\
\hline
\end{tabular}
\caption{\textbf{User study}. Each number represents the proportion of votes for each category.}
\label{tab_huamn_eval_metrics}
\end{table*}

\begin{figure}[!ht]
\centering
\includegraphics[width=0.9\linewidth]{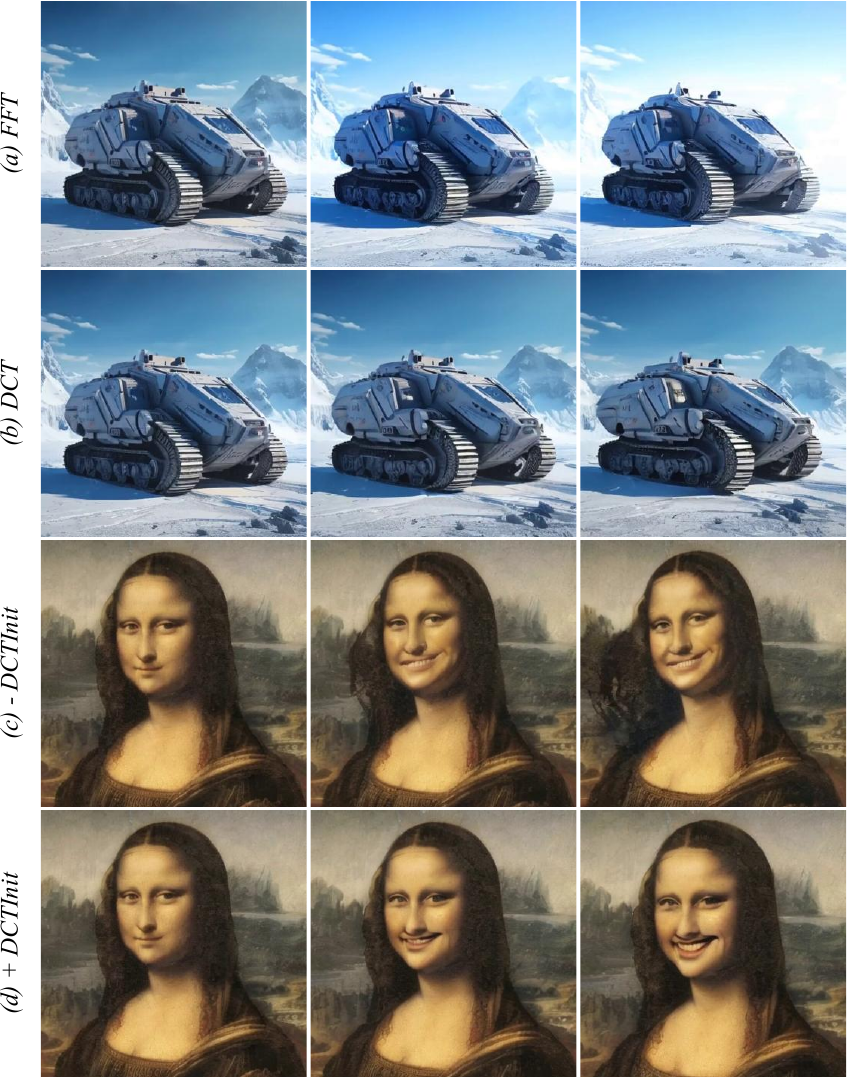}
\caption{(a) vs. (b): The impact of FFT and DCT decomposition. (c) vs. (d): The effectiveness of DCTInit. \textit{Please visit the project page to visualize the animations.}}
\label{fig_fft_vs_dct_e_dctinit}
\end{figure}

\textbf{Image animation visual comparison.} Animated results from ours and our competitors can be found in Fig.~\ref{fig:qualitative_comparison}. For a complete visual comparison, please visit our project page.
The videos generated by PIA significantly deviates from the input image, failing to preserve the fine details of the original image. 
In contrast, I2VGen-XL, DynamiCrafter, and SVD are capable of generating videos with a high degree of motion dynamics, but they struggle to ensure smooth and coherent motion. 
SEINE also tends to drift from the input image and exhibits noticeable distortions in the later frames. 
Meanwhile, ConsistI2V faces challenges in aligning with the textual prompt.
While the commercial tool Hailuo can maintain relatively high consistency with the input image, akin to the performance of PIA, Hailuo tend to loss the details of the given image. Another commercial solution, Genmo, alters the style of the input image, with little consistency with the original input image. 
Our linear attention-based solution, MiraMo, achieves performance comparable to Cinemo while offering a significant advantage in GPU resource efficiency.
All these observations collectively affirm that our model excels in producing coherent and consistent video content in response to specific textual prompts.
Note that we include a disclaimer that the results above are merely a ``snapshot'' of the visual performance of our competitors, as each model can generate multiple results with different seeds. Therefore, this visual comparison should only be considered as a reference and not as conclusive evidence.

\textbf{T2V quantitative comparison.} As shown in Tab.~\ref{tab_t2v_vbench}, our T2V model achieves the highest \textit{average score} while using the smallest model size. Note that we do not compare against large video diffusion models trained with extensive computational resources, such as HunyuanVideo and Wan2.1. Instead, our goal is to demonstrate that our linear attention T2V model can achieve strong generation performance under the limited or similar resources (i.e., computational power and dataset size), highlighting the feasibility of applying linear attention to video generation.

\begin{figure}[h]
\centering
\includegraphics[width=1.0\linewidth]{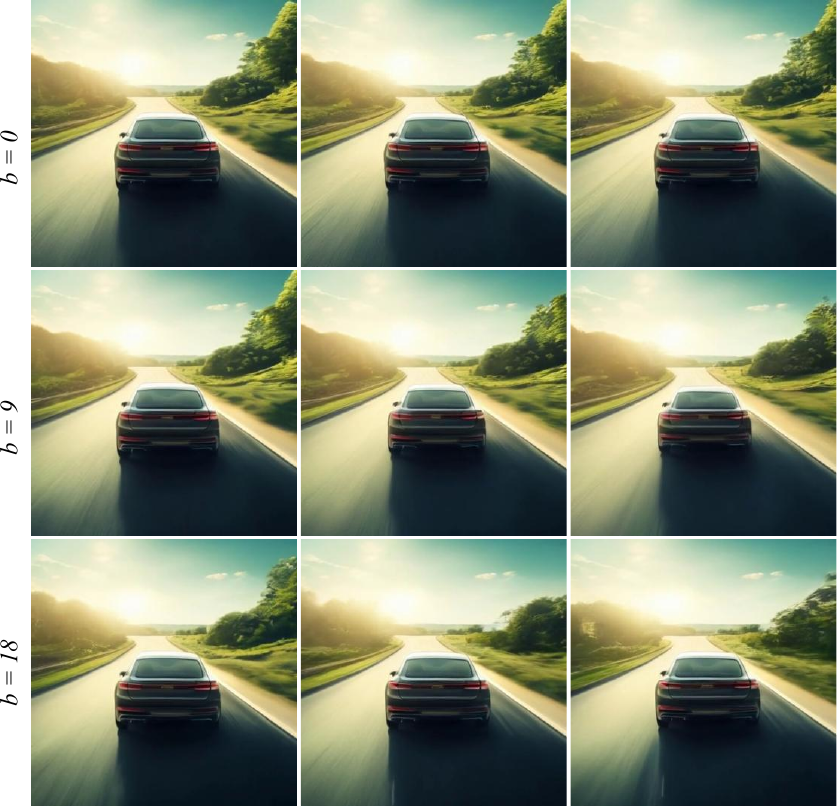}
\caption{{\bf Dynamics degree control.} From top to bottom, the bucket value $b$ increases and effectively increases the speed of the car. \textit{Please visit the project page to visualize the animations.}}
\label{fig_motion_bucket}
\end{figure}

\subsection{Ablation}
\label{experiments_analysis}
\textbf{Motion residuals learning. } 
We first study the effectiveness of motion residual learning (MRL) with a quantitative experiment. To do so, we train a variant of our model without MRL that directly predicts the frames (DPF). The first two rows of Tab.~\ref{table_abalation_components} show the statistics. Motion residual learning effectively improves appearance consistency and smoothness. In both cases, we do not run any noise refinement during the inference time. 

\textbf{DCTInit.}
Next, we evaluate the effect of the noise refinement strategy. Both the FFT and DCT noise refinements are quantitatively evaluated and their statistics are listed in the lower two rows (rows MRL+FFT and MRL+DCT) of Tab.~\ref{table_abalation_components}.  We can see the obvious benefit of enabling DCTInit over the FFT counterpart, as well as the ones (rows DPF and MRL) without any noise refinement strategy.
Fig.~\ref{fig_fft_vs_dct_e_dctinit}(c\&d) further visualizes the difference between results with and without DCTInit via a Mona Lisa example. The result without DCTInit tends to be over-exaggerated. The one with FFT noise refinement even introduces the abrupt color change in the example of  Fig.~\ref{fig_fft_vs_dct_e_dctinit}(a). 

\begin{table}[!t]
\centering
\begin{tabular}{cccc}
\hline
Models          & \begin{tabular}[c]{@{}c@{}}I2V Average \\ Consistency\end{tabular} & \begin{tabular}[c]{@{}c@{}}Average \\ Consistency\end{tabular} & \begin{tabular}[c]{@{}c@{}}Dynamic \\ Degree\end{tabular} \\ \hline
DPF & 92.62 & 93.62 & 32.68 \\
MRL & 94.02 & 94.53 & 35.12 \\
MRL+FFT & 94.93 & 95.02 & 33.90 \\
MRL+DCT & 95.47 & 95.14 & 35.93 \\ \hline
\end{tabular}
\caption{DPF stands for Direct Predict Frames, while MRL refers to Motion Residuals Learning. Our proposed method is represented in the final row (MRL+DCT). “Average” refers to the mean of subject consistency and background consistency.}
\label{table_abalation_components}
\vspace{-0.3cm}
\end{table}

\textbf{Dynamics degree control.} We assess the dynamic degree using four metrics: optical flow, mean absolute difference, MS-SSIM, and SSIM across frames. Our study begins by randomly selecting 1,000 videos from the training dataset. For each video, we extract a 16-frame segment starting from the first frame, using different frame intervals. The frame intervals used are 3, 5, 7, 9, 11, 13, 15, 17, 19, 21, 23, and 25. Generally, videos with a frame interval of 3 tend to exhibit lower overall motion intensity compared to those with a frame interval of 7, and so on.

\begin{figure}[!ht]  
\centering
\subfloat[Mean absolute difference]
  {
     \includegraphics[width=0.45\linewidth]{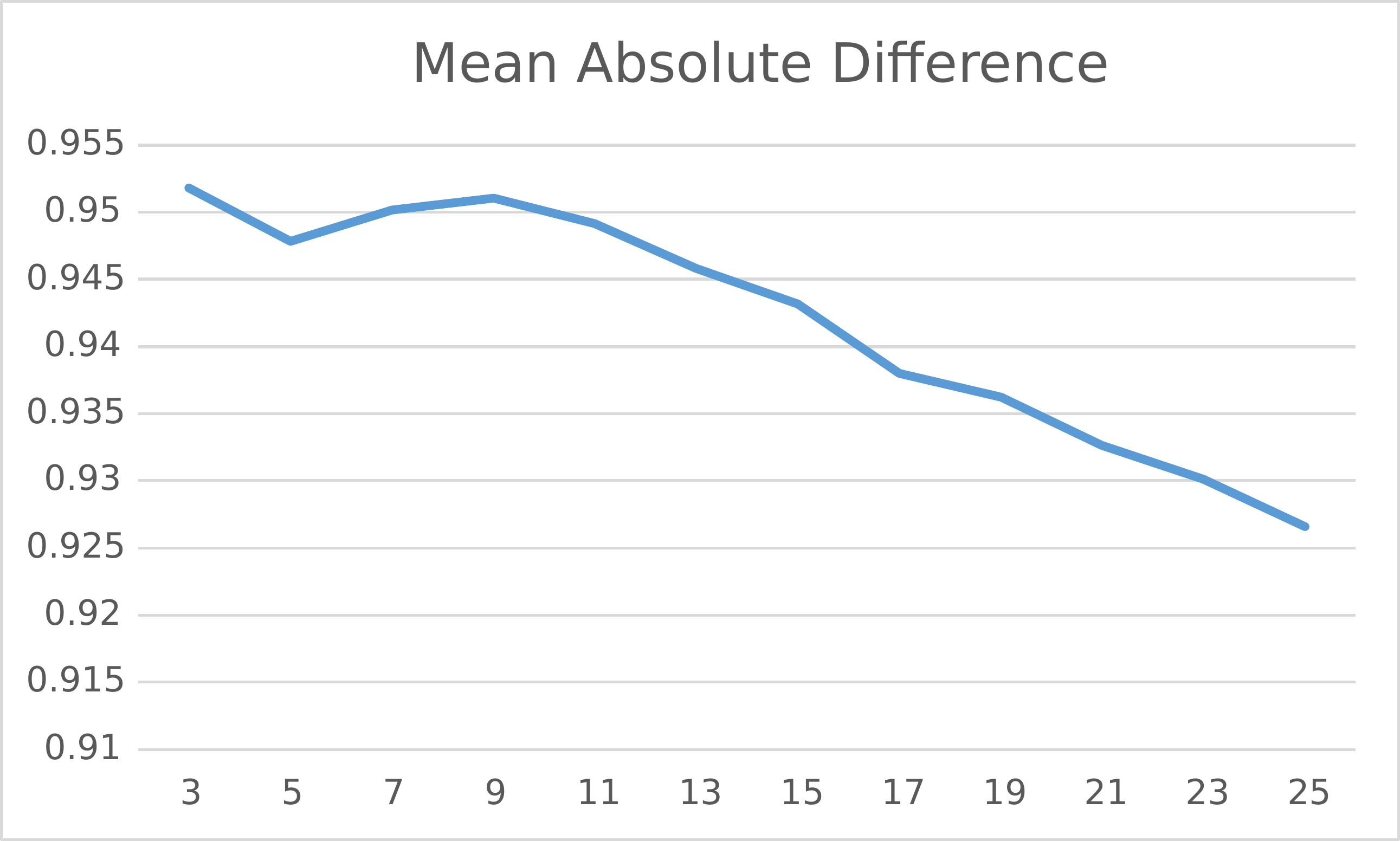}
  }
\subfloat[Optical flow]
  {
     \includegraphics[width=0.45\linewidth]{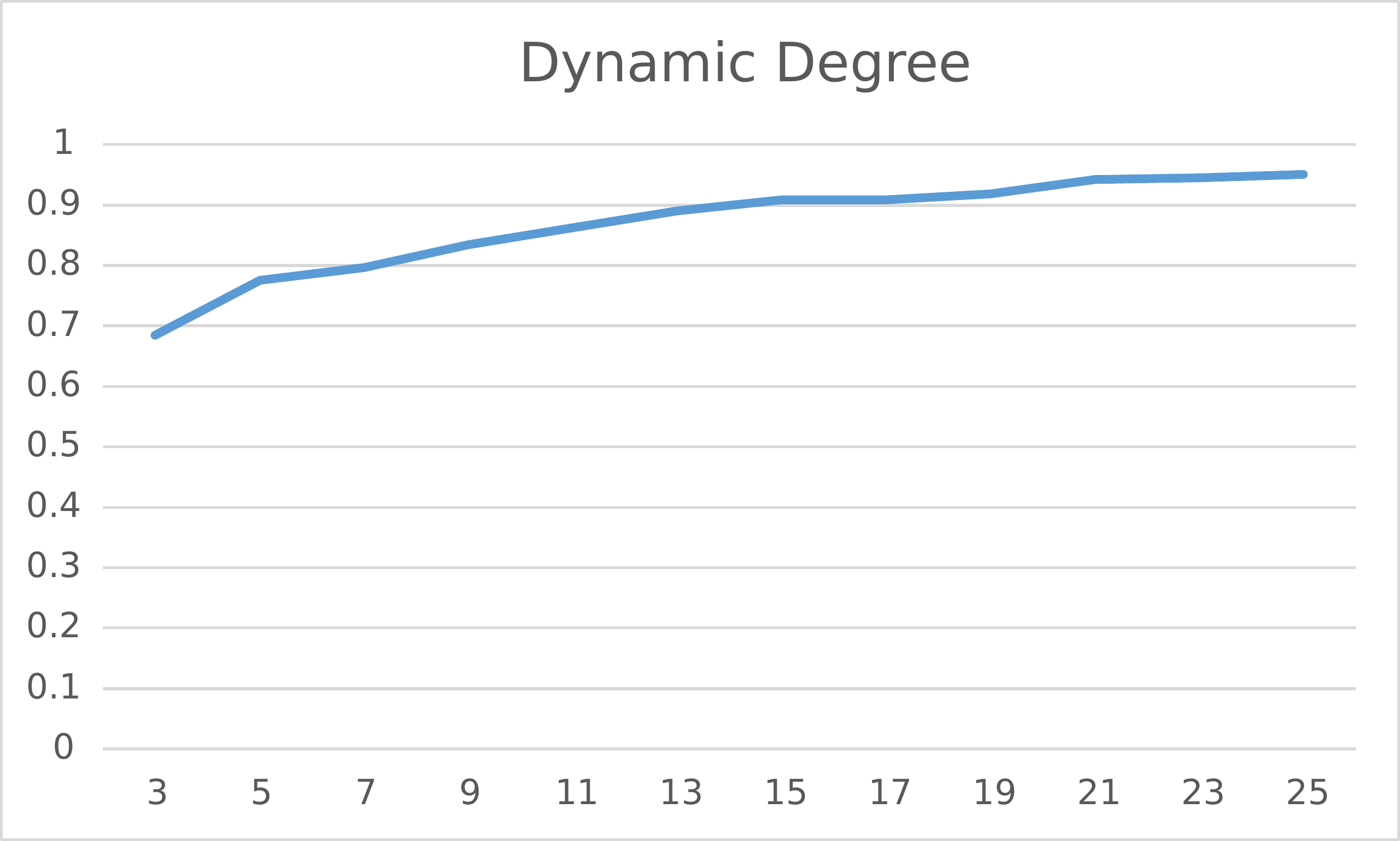}
  } \\
  
  \subfloat[SSIM]
  {
     \includegraphics[width=0.45\linewidth]{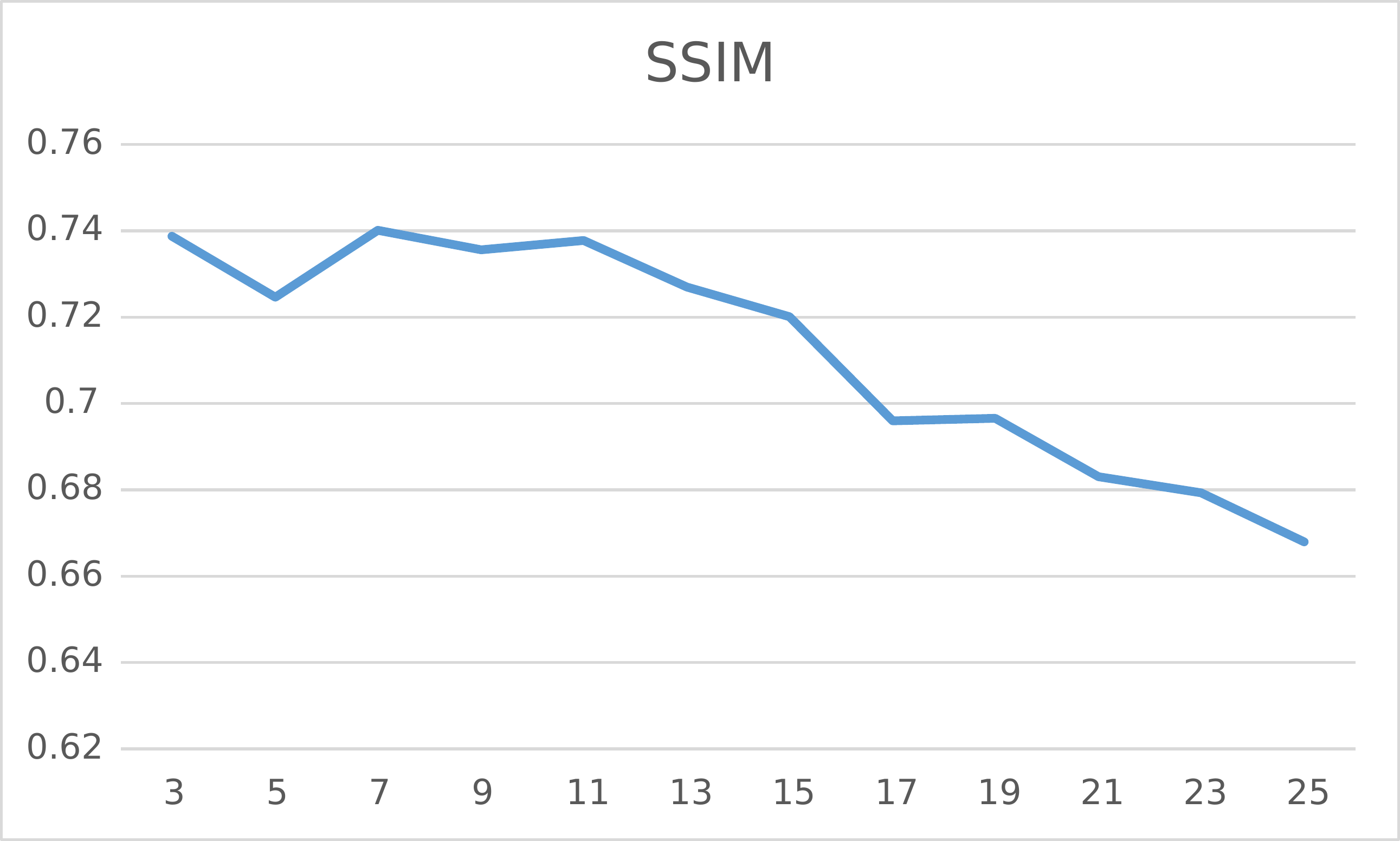}
  }
\subfloat[MS-SSIM]
  {
    \includegraphics[width=0.45\linewidth]{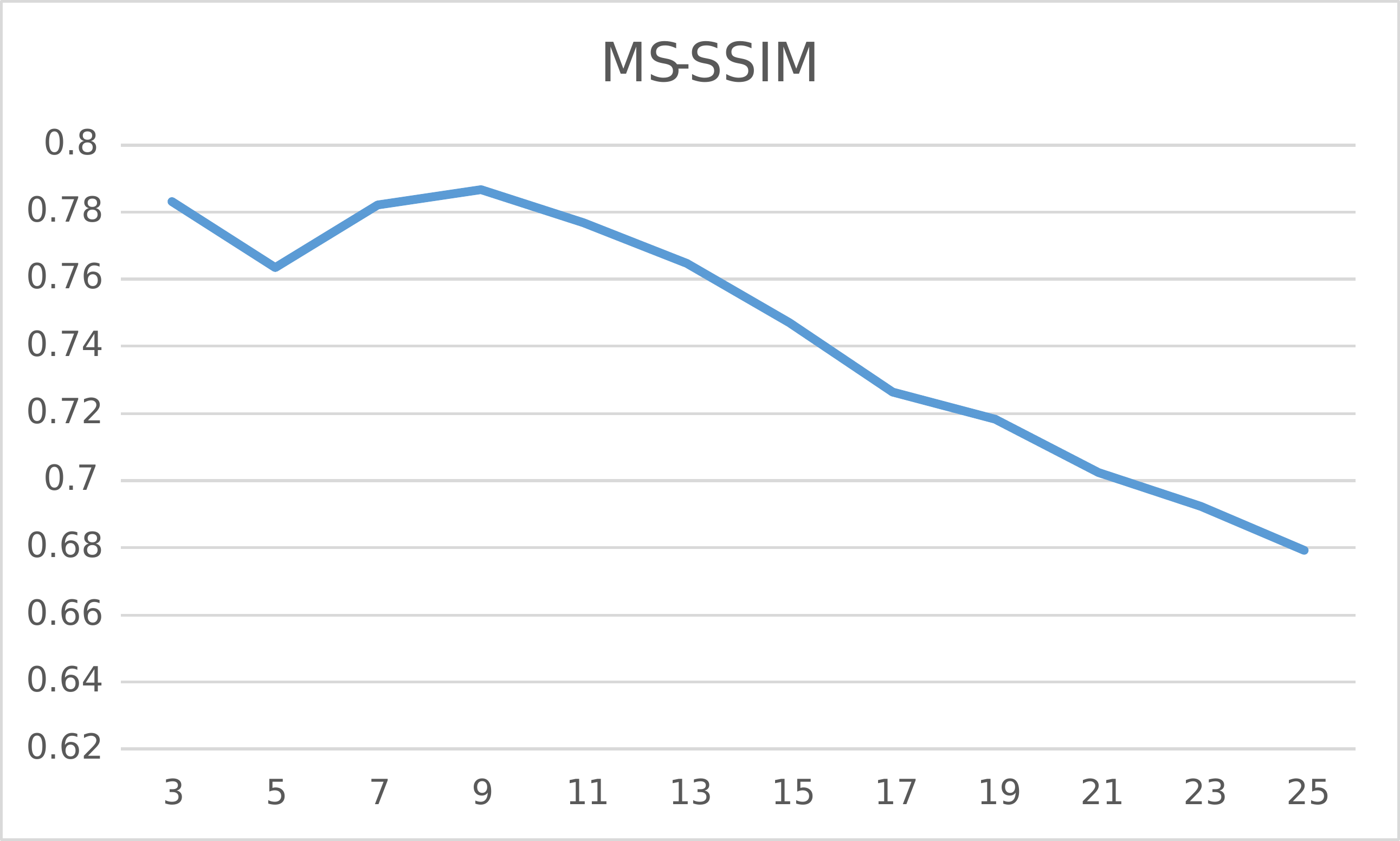}
  } 

\caption{The performance differences between four different motion dynamic estimation methods.}
\label{fig_motion_intensity_methods}
\end{figure}

Subplots (a), (c), and (d) in Fig.~\ref{fig_motion_intensity_methods} illustrate the overall motion dynamics degree of the video groups estimated using the mean absolute difference between frames, the mean structural similarity index (SSIM), and the mean multi-scale structural similarity index (MS-SSIM), respectively. The vertical axis in these subplots represents the similarity of the videos, and theoretically, the similarity should gradually decrease as the frame interval increases. Subplot (b) shows the dynamic degree estimated using the RAFT optical flow estimator~\cite{teed2020raft}, with the vertical axis representing the dynamic degree. Theoretically, this value should increase as the frame interval grows.

As shown in Fig.~\ref{fig_motion_intensity_methods}, the mean absolute difference between frames, the mean MS-SSIM, and the dynamic degree estimated through optical flow can all accurately reflect the motion intensity of the videos. However, in terms of computational efficiency, the average time required to estimate the motion intensity of a video using these three methods is 0.03 seconds, 0.07 seconds, and 1.17 seconds, respectively. Given this, we exclude the optical flow-based method. Additionally, considering that the MS-SSIM-based method more accurately reflects changes in motion as perceived by human vision, while the mean absolute difference method is overly sensitive to noise or changes in non-essential details (for example, for videos with high color contrast but low actual motion intensity, the mean absolute difference might be exaggerated), we ultimately choose the MS-SSIM-based method to estimate the motion intensity of the videos.

The degree of dynamics of the generated videos can be controlled via the bucket $b$ as mentioned above. Fig.~\ref{fig_motion_bucket} visualizes how the motion changes when we gradually increase the bucket value $b$ from 0 to 18, through the  example ``{\tt\small car moving}.'' As $b$ increases in value, the driving distance of the car increases significantly.

\begin{figure}[!ht]
\centering
\includegraphics[width=1.0\linewidth]{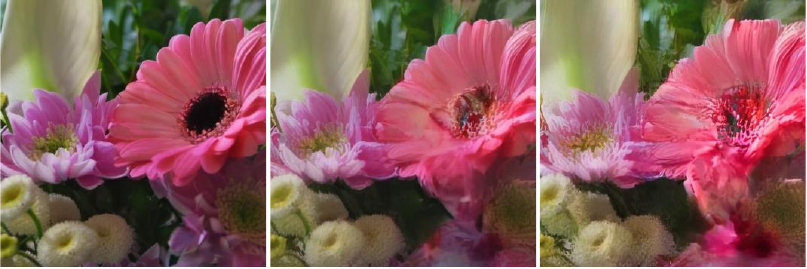}
\caption{\textbf{Results by using alternative motion residual representation.} The generated videos tend to degrade rapidly.}
\label{fig_motion_representaion_2}
\end{figure}

\textbf{Alternative motion residual representation.} In Sec.~\ref{method_image_animation_formulation}, we obtain motion residuals using the equation $\mathbf{M}=\{z_2-z_1,z_3-z_1,\ldots,z_N-z_1\}$. Inspired by common practices in video compression, we introduce a different representation of motion residuals using the equation $\mathbf{M}=\{z_2-z_1,z_3-z_2,\ldots,z_N-z_{N-1}\}$. As shown in Fig~\ref{fig_motion_representaion_2}, we observe that using this representation of motion residuals results in rapid degradation in the generated videos. We think this phenomenon is primarily caused by error accumulation.

\textbf{The benefit of synthetic video data.}
\label{sec_the_benefit_of_synthetic_video_data}
As shown in Tab.~\ref{table_abalation_real_vs_sync}, the synthetic video dataset enhances the quality of the generated videos. The model trained on synthetic video data achieves over a 3\% improvement in average VBench scores compared to the model trained on real video data.

\begin{table}[!ht]
\centering
\begin{tabular}{ccccc}
\hline
Models & \begin{tabular}[c]{@{}c@{}}Multiple \\ Object\end{tabular} & \begin{tabular}[c]{@{}c@{}}Human \\ Action\end{tabular} & Color & \begin{tabular}[c]{@{}c@{}}Spatial \\ Relationship\end{tabular} \\ \hline
Real data & 52.91 & 92.00 & 90.55 & 37.18 \\
Sync data & 45.50 & 88.40 & 88.40 & 35.50 \\ \hline
\end{tabular}
\caption{Performance differences across different post-training datasets.}
\label{table_abalation_real_vs_sync}
\end{table}

\subsection{Discussions and Further Applications}
\label{sec_discussions_and_further_applications}
\begin{figure}[!ht]
\centering
\includegraphics[width=1.0\linewidth]{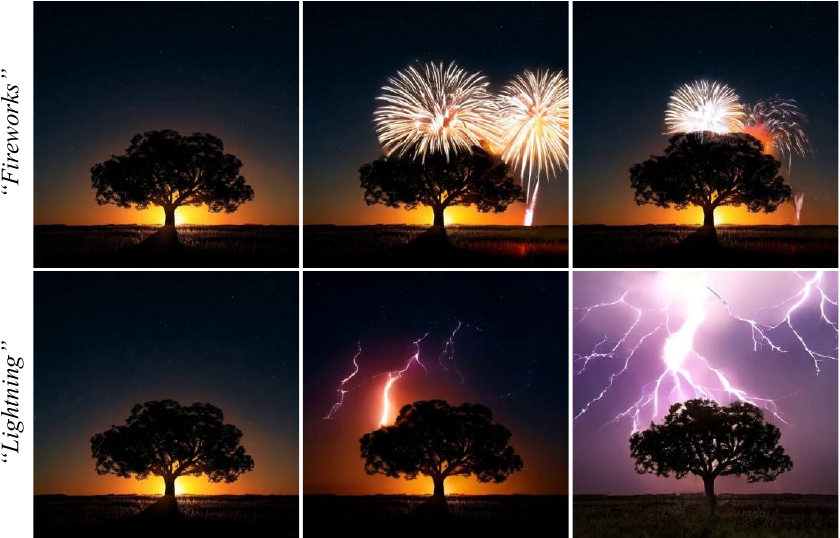}
\caption{\textbf{Motion control by textual prompts.} Our model can effectively respond to textual prompts, leading to visually appealing outcomes. \textit{Please visit the project page to visualize the animations.}}
\label{fig_motion_control}
\end{figure}

\textbf{Motion control by prompting.} By learning the distribution of motion residuals rather than directly predicting frames, our design strengthens the connection between the predicted motion residuals and the static input image, as well as significantly improves the alignment accuracy between the video content and the user text prompts.
As demonstrated in Fig.~\ref{fig_motion_control}, our model does not rely on complex guiding instructions, and can accurately respond to the text prompts from users.

\begin{figure}[!t]
\centering
\includegraphics[width=1.0\linewidth]{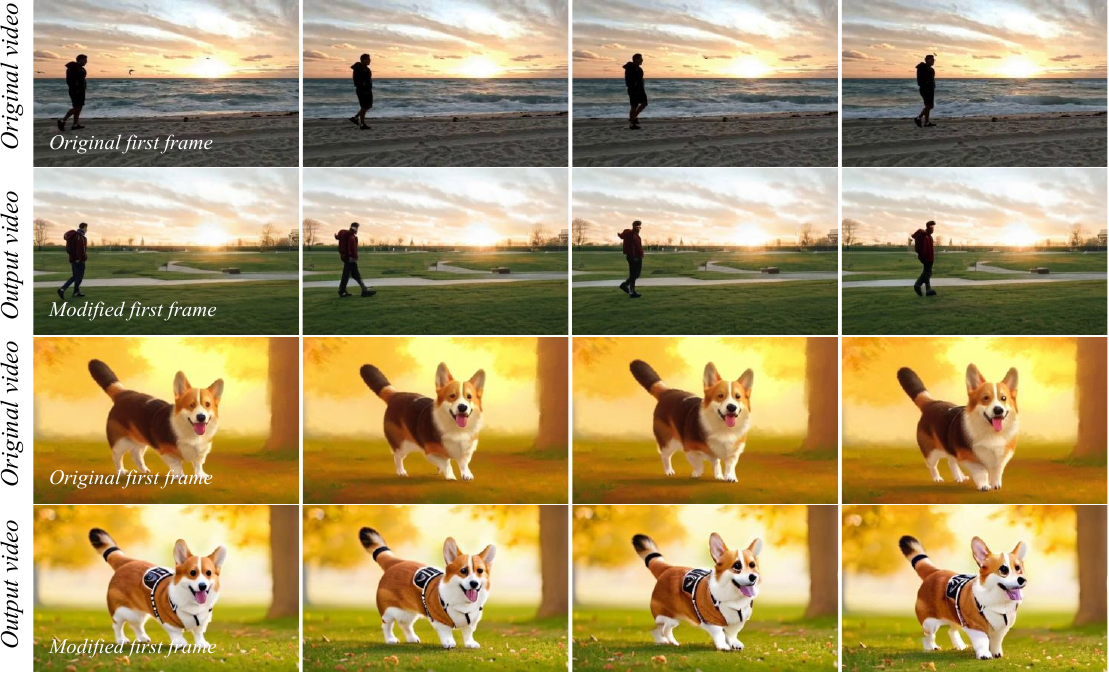}
\caption{\textbf{Motion transfer/Video editing results.} Our model can be easily extended to motion transfer or video editing on any given video from any source, not necessarily the ones generated by our model. Our model can also handle different resolutions when applied to other applications. \textit{Please visit the project page to visualize the animations.}}
\label{fig_motion_transfer_video_editing}
\end{figure}

\textbf{Motion transfer/Video editing.} The advantage of learning motion residuals is that it allows us to achieve motion transfer of the given video to a new one. 
Note that the provided video can originate from any source, including videos generated by video generative models or real-world footage.
To do so, we first perform the rectified flow inversion~\cite{avrahami2025stable} to obtain the initial inference noise corresponding to the motion residuals of the given video.
Next, the first frame of the video can be modified with any off-the-shelf image editing tools (e.g. Paintbrush) or more sophisticated techniques~\cite{tumanyan2023plug}. 
Finally, the motion-transferred video can be generated by feeding the extracted inference noise and the modified first frame to our model. 
Fig.~\ref{fig_motion_transfer_video_editing} demonstrates two examples of motion transfer or video editing.

\textbf{Limitations.} We use an image-level VAE to compress the training videos, which may result in flickering in the generated videos. To mitigate this, we plan to introduce a temporal video VAE. In addition, we only discusses the training of SFT, and in the future, we will introduce DPO (Direct Preference Optimization) to enhance the quality of the generated videos further.

%% file: sections/conclusion.tex
\section{Conclusion}
In this paper, we introduce MiraMo, a novel image animation model built on a linear attention Transformer. MiraMo achieves improved image consistency and smoother motion while offering greater GPU efficiency and higher throughput. Our key idea of image animation is to focus on learning the distribution of motion residuals rather than directly predicting the next frames. During the inference, we propose DCTInit, which utilizes the low-frequency components of DCT coefficients of the static input image to refine initial noise and stabilize the generation process. Our experimental results demonstrate the effectiveness and superiority of our approach compared to existing methods.